\documentclass{article}

    \usepackage[final]{neurips_2025}

\usepackage[utf8]{inputenc} 
\usepackage[T1]{fontenc}    
\usepackage{hyperref}       
\usepackage{url}            
\usepackage{booktabs}       
\usepackage{amsfonts}       
\usepackage{nicefrac}       
\usepackage{microtype}      
\usepackage{xcolor, amsmath}         
\usepackage{latexsym}
\usepackage{appendix, subfigure, caption, picinpar}
\usepackage{wrapfig}
\usepackage{graphicx}
\usepackage{tcolorbox, multicol, multirow, makecell}
\usepackage{algorithm}
\usepackage{etoc}
\usepackage{lipsum}
\tcbuselibrary{skins, breakable, theorems}

\usepackage{booktabs} 
\usepackage{xcolor}   
\usepackage{colortbl} 
\usepackage{algpseudocode} 
\usepackage{amsmath} 
 
\usepackage{amssymb} 
\usepackage{multirow}
\usepackage{amsfonts}
\usepackage{wrapfig}
\usepackage{subcaption}
\usepackage{enumitem}
\title{UtilGen: Utility-Centric Generative Data Augmentation with Dual-Level Task Adaptation}

\author{
Jiyu Guo$^{1}$ \quad
Shuo Yang$^{1}$\thanks{Corresponding author.} \quad
Yiming Huang$^{1}$ \quad
Yancheng Long$^{1}$ \quad
Xiaobo Xia$^{2,3}$ \quad \\
\textbf{
Xiu Su$^{4}$ \quad
Bo Zhao$^{5}$ \quad
Zeke Xie$^{6}$ \quad
Liqiang Nie$^{1}$
} \\
$^{1}$Harbin Institute of Technology, Shenzhen \quad
$^{2}$National University of Singapore \\
$^{3}$MoE Key Laboratory of Brain-inspired Intelligent Perception and Cognition, \\ University of Science and Technology of China \\
$^{4}$Central South University \quad
$^{5}$Shanghai Jiao Tong University \\
$^{6}$Hong Kong University of Science and Technology (Guangzhou) \\
\{ \texttt{220110126@stu.hit.edu.cn, shuoyang@hit.edu.cn} \}
}
  
\begin{document}

\maketitle

\begin{abstract}

Data augmentation using generative models has emerged as a powerful paradigm for enhancing performance in computer vision tasks. However, most existing augmentation approaches primarily focus on optimizing intrinsic data attributes -- such as fidelity and diversity -- to generate visually high-quality synthetic data, while often neglecting task-specific requirements. Yet, it is essential for data generators to account for the needs of downstream tasks, as training data requirements can vary significantly across different tasks and network architectures.
To address these limitations, we propose \textsc{UtilGen}, a novel utility-centric data augmentation framework that adaptively optimizes the data generation process to produce task-specific, high-utility training data via downstream task feedback. Specifically, we first introduce a weight allocation network to evaluate the task-specific utility of each synthetic sample. Guided by these evaluations, \textsc{UtilGen} iteratively refines the data generation process using a dual-level optimization strategy to maximize the synthetic data utility: (1) model-level optimization tailors the generative model to the downstream task, and (2) instance-level optimization adjusts generation policies -- such as prompt embeddings and initial noise -- at each generation round.
Extensive experiments on eight benchmark datasets of varying complexity and granularity demonstrate that \textsc{UtilGen} consistently achieves superior performance, with an average accuracy improvement of 3.87\% over previous SOTA. Further analysis of data influence and distribution reveals that \textsc{UtilGen} produces more impactful and task-relevant synthetic data, validating the effectiveness of the paradigm shift from visual characteristics-centric to task utility-centric data augmentation.

\end{abstract}

\section{Introduction}

\label{intro}
Recent advances in generative models, particularly diffusion models \cite{ho2020denoising, dhariwal2021diffusion, wang2025lavin, rombach2022high, ho2022classifier, luo2024deem, song2020denoising}, have significantly advanced data augmentation by enabling the creation of photorealistic images. Such text-to-image systems are capable of generating diverse and high-fidelity samples, and empirical evidence has shown their potential to enhance downstream model performance \cite{he2022synthetic}. 

Current generative data augmentation approaches can be categorized into two main paradigms: \textit{fidelity preservation} and \textit{diversity enhancement}. The former employs techniques such as LoRA-based fine-tuning \cite{hu2022lora} to align synthetic data with real-world distributions \cite{yuan2023real, kim2024datadream}, while the latter employs varied prompts or feature perturbations to enhance data diversity \cite{zhang2023expanding, da2023diversified}. Although effective in generating visually high-quality data, these methods solely optimize the intrinsic data attributes (e.g., fidelity and diversity), often struggling to directly optimize the task-specific utility of synthetic data. In practice, different tasks and model architectures may require distinct training data distributions for optimal performance \cite{zha2025data}, as exemplified in Figure~\ref{fig:1}. Despite this, most existing methods lack mechanisms to adapt data generation process based on the needs of specific downstream tasks. This limitation motivates our investigation into utility-centric data augmentation, in which synthetic data is explicitly optimized to enhance task performance rather than merely meet visual standards.

To go beyond the above limitation, an effective mechanism is needed to assess the task-specific utility of synthetic data, thereby providing explicit optimization signals to guide the data augmentation process. However, evaluating utility through full training and testing cycles is computationally prohibitive. Therefore, the core challenges in developing utility-centric data augmentation are: (1) \textit{How to efficiently evaluate the task-specific utility of synthetic data without exhaustive training?} and (2) \textit{How to systematically improve the task-specific utility of synthetic data?}

In this work, we propose \textsc{UtilGen}, a novel utility-centric data augmentation framework, which can adaptively optimize the data generation process to produce task-specific, high-utility training data based on downstream task feedback. Specifically, we introduce \textit{Task-Oriented Data Valuation}, which quantifies the task-specific utility of synthetic data through a meta-learned weight allocation network \cite{shu2019meta, shu2023cmw, zhang2021learning}. The network is optimized to minimize the classifier's validation loss by adaptively weighting the losses of training samples via estimation of their utility.
 The trained valuation network serves as an efficient utility predictor, enabling assessment of task-specific utility for newly generated samples without the need for costly retraining cycles. Guided by the utility signals, we employ an integrated dual-level optimization strategy: (1) \textit{Model-Level Generation Capability Optimization} that tailors the data generator to downstream tasks through Direct Preference Optimization (DPO), and (2) \textit{Instance-Level Generation Policy Optimization} that optimizes the generation policies (\textit{i.e.}, prompt embedding and initial noise) to maximize the task-specific utility of synthetic data. Compared to previous advanced data augmentation methods which focus on optimizing intrinsic data characteristics, our proposed method achieves an average accuracy improvement of 3.87\% across eight benchmark datasets. To the best of our knowledge, this is the first generative augmentation method where ResNet-50 \cite{he2016deep} trained solely on 3$\times$ synthetic data surpasses its real-data-trained counterpart on several benchmarks. Before delving into the details, we summarize our contributions as follows:

\begin{figure*}[t]
    \centering
    \includegraphics[width=\textwidth]{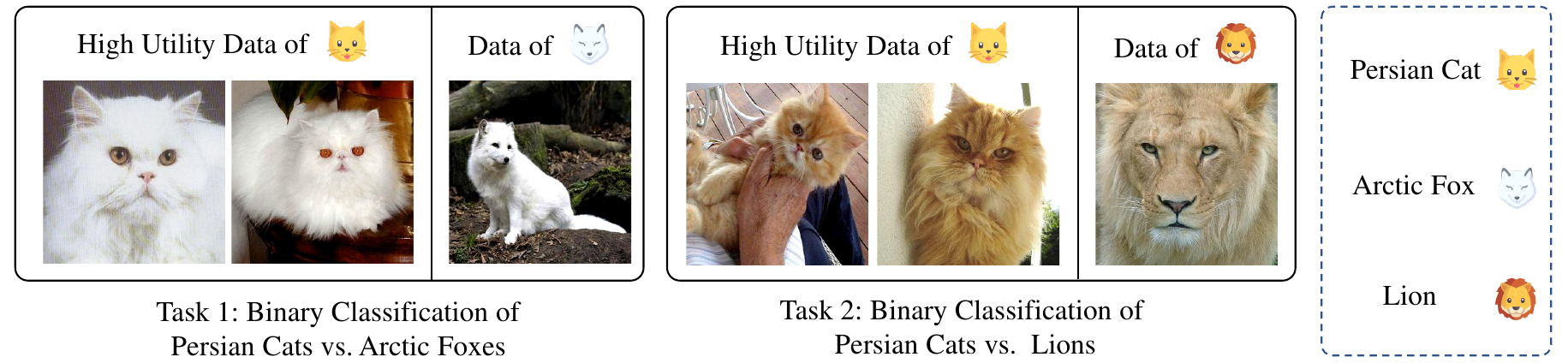}
    
    \caption{Comparison of high-utility samples within the same category (Persian cats) across two different tasks. White Persian cats (left) are more useful in Task 1, while golden ones (right) are more beneficial in Task 2, highlighting the diverse data requirements in different downstream tasks.  }
    \label{fig:1} 
    \vspace{-0.7em}
\end{figure*}

\begin{itemize}[leftmargin=1.7em]
\item Motivated by the observation that training data requirements differ across tasks and network architectures, we introduce a novel paradigm shift in data augmentation. Instead of focusing solely on optimizing intrinsic data attributes, we emphasize enhancing the task-specific utility of synthetic data. This utility-centric approach adaptively optimizes the generation process according to downstream task needs, enabling more targeted and effective data augmentation.

\item To efficiently evaluate the task-specific utility of synthetic data, we introduce a meta-learned weight allocation network that measures the utility of synthetic data without requiring costly retraining. These utility signals drive a dual-level optimization framework that enhances both the model generation capability and the generation policies, resulting in high-utility synthetic data tailored to downstream tasks.

\item Our method achieves state-of-the-art performance with an average improvement of 3.87\% in accuracy across eight benchmarks, while also demonstrating exceptional versatility by delivering consistent performance gains across diverse architectures (e.g., ResNeXt\cite{xie2017aggregated}, WideResNet \cite{zagoruyko2016wide}, and MobileNet\cite{sandler2018mobilenetv2}). Through training trajectory analysis, we validate the rationality of task utility measurement based on the weight network. Furthermore, analyses of data influence and distribution reveals that \textsc{UtilGen} generates data with higher task relevance and stronger positive impact on model performance.
\end{itemize}

\section{Related Work}

\subsection{Training Data Valuation}

Understanding the role of training data in model performance is crucial for data-efficient learning \cite{zhao2020dataset, zhao2023dataset, cao2023data, yang2024mind, yang2023bicro} and optimizing model behavior \cite{zhang2025benchmark, cao2024mentored, huang2025diffusion, zhang2021deep, yang2022estimating}. To better understand how training data affects model behavior, many studies have aimed to quantitatively assess the influence of individual examples on model performance \cite{koh2017understanding, ghorbani2019data, kwon2021efficient, jiang2020characterizing, nohyun2022data,wu2022davinz,zhang2022rethinking, yang2022dataset}. Existing approaches for evaluating data value can be classified into two categories:  (1) Retraining-based approaches, such as Data Shapley \cite{ghorbani2019data, kwon2021efficient, grabisch1999axiomatic, bordt2023shapley} and C-score \cite{jiang2020characterizing}, which quantify data influence through expensive model retraining across different training subsets. For the utility evaluation of large-scale synthetic datasets, these methods become computationally prohibitive due to their inherent complexity. (2) Gradient-based methods \cite{koh2017understanding, yeh2018representer, pruthi2020estimating, chen2021hydra} that estimate data influence by analyzing gradient interactions between training and test points, either through static snapshot analysis or dynamic trajectory examination. Although these approaches avoid model retraining, they still incur significant computational overhead, particularly when performing complex operations such as Hessian matrix inversion \cite{hammoudeh2024training}. To evaluate the utility of synthetic data, our method employs a weight allocation network \cite{shu2019meta, shu2023cmw, zhang2021learning} to efficiently assess data utility, avoiding the costly retraining or complex computations required by earlier approaches.

\subsection{Training Data Augmentation}

The availability of high-quality training data has been fundamental to the success of deep learning, enabling models to capture complex patterns, learn meaningful representations, and generate accurate predictions \cite{zhu2016we, zhang2025logosp, liu2025trackvla++, zhou2024few, zhang2025visible}. The methodology of training data augmentation has progressed from traditional techniques to advanced generative approaches \cite{huang2022harnessing, yuan2023real, kim2024datadream, dunlap2023diversify, zhang2023expanding, liao2025bood, yang2021free}. Traditional methods, such as mixup \cite{hendrycks2019augmix, zhang2020does}, erasing \cite{zhong2020random, devries2017improved}, and cropping \cite{ciocca2007self}, commonly rely on predefined transformations to augment dataset diversity. However, they are inherently limited to local pixel-level modifications. While Generative Adversarial Networks (GANs)~\cite{goodfellow2014generative} enabled synthetic image generation, they often face challenges in maintaining semantic consistency and distribution alignment ~\cite{frid2018synthetic,zhu2017unpaired}. Recent advances in diffusion models, such as Stable Diffusion~\cite{rombach2022high} and GLIDE~\cite{nichol2021glide}, have demonstrated superior capabilities in generating synthetic data. To enhance diversity, methods such as GIF \cite{zhang2023expanding}, ALIA \cite{dunlap2023diversify}, and DISEF \cite{da2023diversified} employ varied prompts or feature perturbations in the latent space. Meanwhile, techniques like RealFake \cite{yuan2023real}, DistDiff \cite{zhu2024distribution}, and DataDream \cite{kim2024datadream} focus on improving image fidelity by aligning synthetic data distributions with the target domain. Nevertheless, these methods primarily address intra-class distribution alignment without evaluating which types of data better support downstream tasks. In contrast, our approach adaptively tailors the generation process to downstream tasks, producing high-utility data specifically optimized for target applications.

\section{Method}

\begin{figure*}[t]
    \centering
    \includegraphics[width=\textwidth]{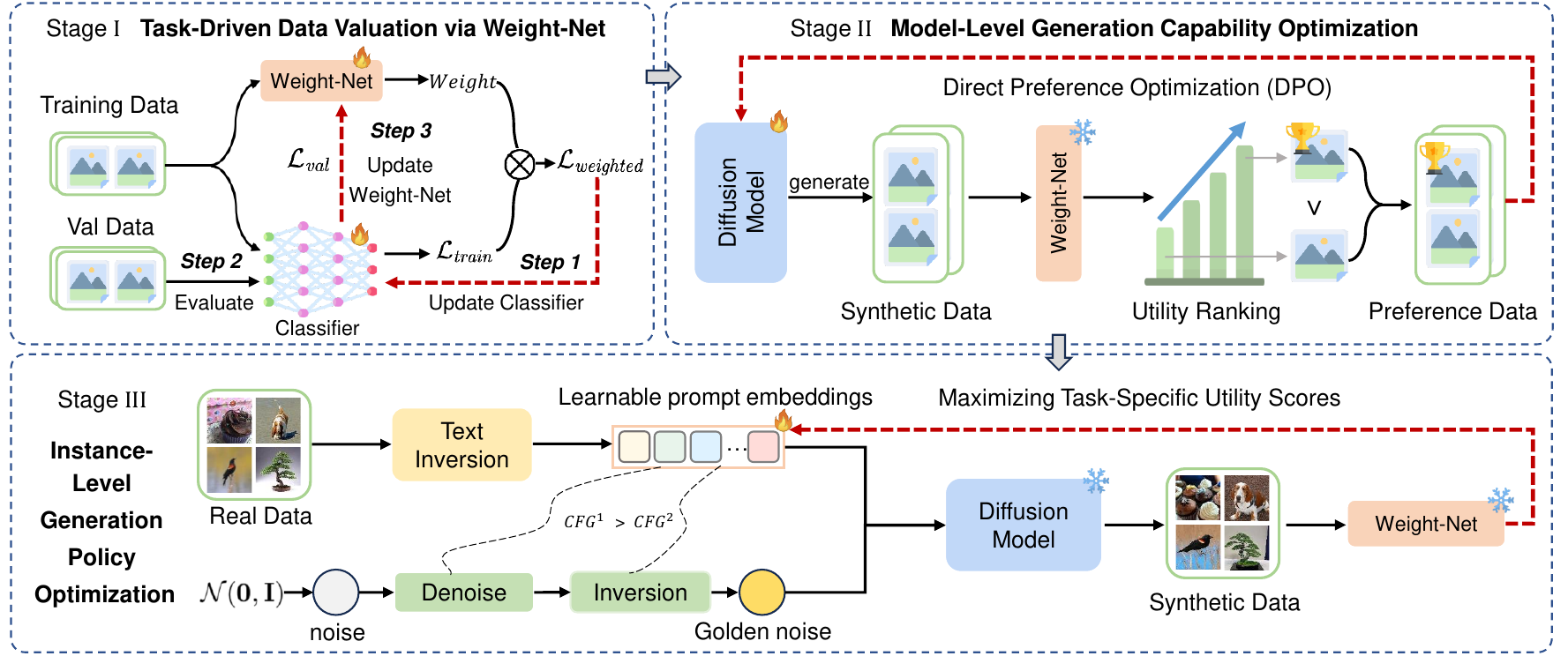}
    \caption{The \textsc{UtilGen} framework for feedback-driven data augmentation, comprising three key stages: (1) \textit{Task-Oriented Data Valuation} (Sec.~\ref{sec:todv}); (2) \textit{Model-Level Generation Capability Optimization} (Sec.~\ref{sec:mlpo}); (3) \textit{Instance-Level Generation Policy Optimization} (Sec.~\ref{sec:ilo}).}
    \label{fig:framework} 
    \vspace{-0.7em}
\end{figure*}

In this section, the framework of our utility-centric generative data augmentation system is presented, as illustrated in Fig.~\ref{fig:framework}. Specifically, the proposed approach comprises three key components: (1) \textit{Task-Oriented Data Valuation (Sec.~\ref{sec:todv})}, which quantitatively assesses the utility of synthetic data for downstream tasks; (2) \textit{Model-Level Generation Capability Optimization (Sec.~\ref{sec:mlpo})}, which tailors the generative model to align with the training data preferences of the downstream task via DPO; and (3) \textit{Instance-Level Generation Policy Optimization (Sec.~\ref{sec:ilo})}, which adapts generation policy to maximize the task-specific utility of the synthetic data.

\subsection{Task-Oriented Data Valuation (TODV)}
\label{sec:todv}

It is noted that individual training instances exhibit heterogeneous influences on model performance during the training process~\cite{koh2017understanding}, with certain data samples potentially introducing negative influences. This observation has motivated a line of research to mitigate model overfitting to data samples with negative influences, such as training sample re-weighting strategies implemented through learnable weight networks\cite{shu2019meta, shu2023cmw, zhang2021learning}. Moreover, it is recognized that these learned weights can be implicitly interpreted as indicators of each sample's utility for downstream tasks~\cite{shu2019meta}. Drawing inspiration from this insight, we propose \textit{Task-Oriented Data Valuation}, which employs a weight network trained via meta-learning to quantitatively assess the utility of synthetic data for downstream tasks.

Specially, given a classifier $ f_\theta $, the weight $ \omega_i $ assigned to each sample $ x_i $ is derived through a loss-based process as follows:
\begin{equation}
\omega_i = \mathcal{W}_\phi \left( \mathcal{L}(f(x_i;\theta), y_i) \right),
\label{eq:weight}
\end{equation}
where $ \mathcal{W}_\phi $ denotes an MLP network with a single hidden layer. This network is trained to predict normalized weights within the range $[0, 1]$, where higher values reflect samples with greater utility.

To develop such a weight network \( \mathcal{W}_\phi \) capable of measuring task-specific data utility, we adopt a bi-level optimization strategy comprising two iterative steps:


\textbf{Classifier training: } To enhance the generalization capability of the weight network and mitigate distribution shift in subsequent generation optimization stages, we first perform textual inversion~\cite{gal2022image} to learn class-specific identifiers $[I_i]$ from a small set of real images per class. Using these learned identifiers, we generate synthetic data $\mathcal{D}_g$ with class-conditioned prompts $c_i$ of the form ``a photo of $[I_i]$''. The synthetic data is then combined with the real training data to form a merged dataset $\mathcal{D}_{\text{merge}} = (\mathcal{D}_r \cup \mathcal{D}_g)$, where $\mathcal{D}_r$ denotes the real data. The classifier parameters $\theta$ are optimized using a \textit{weighted loss} computed over $\mathcal{D}_{\text{merge}} = \{(x_i, y_i)\}_{i=1}^N$:
\begin{equation}
    \theta^*(\phi) = \arg\min_{\theta} \frac{1}{N} \sum_{i=1}^N \omega_i \mathcal{L}(f(x_i;\theta), y_i),
\label{eq:classifier_update}
\end{equation}

\textbf{Weight network training: } Given the classifier's updated parameters, the weight network parameters \( \phi \) are trained to minimize the loss on the validation set \( \mathcal{D}_v = \{(x_j, y_j)\}_{j=1}^M \):

\begin{equation}
    \phi^*(\theta)=\arg\min_{\phi}\frac{1}{M} \sum_{j=1}^M \mathcal{L}\left(f(x_j;\theta^*(\phi)), y_j\right).
\label{eq:weight_network_update}
\end{equation}

This bi-level optimization framework establishes a dynamic feedback loop between data valuation and model training. By quantifying data utility through learned weights, the trained weight network is subsequently used to guide the optimization of the data generation process. The full procedure for TODV is described in Algorithm \ref{algo_todv}.

\begin{algorithm}
\caption{Task-Oriented Data Valuation (TODV)}\label{algo_todv}
\vspace{-0.3em}
\footnotesize
\begin{flushleft}
\textbf{Input}: Training data $\mathcal{D}_r \cup \mathcal{D}_g = \{(x_i, y_i)\}_{i=1}^N$; validation data $\mathcal{D}_v = \{(x_j, y_j)\}_{j=1}^M$

\textbf{Required}: Classifier parameters $\theta$ ; weight network parameters $\phi$;  max iteration $T$. 
\end{flushleft}
\begin{algorithmic}[1]
\State Initialize $\theta^{(0)}$, $\phi^{(0)}$ randomly
\For{t = 0 to $T-1$}
    \State Sample batch $\{(x_i, y_i)\}_{i=1}^n$ from $\mathcal{D}_r \cup \mathcal{D}_g$
    \State Sample batch $\{(x_j, y_j)\}_{j=1}^m$ from $\mathcal{D}_v$
    \State Predict weights $\{\omega_i^{(t)}\}_{i=1}^n$ for the batch $\{(x_i, y_i)\}_{i=1}^n$ that reflect their task-specific utility (Eq. \ref{eq:weight})
    \State Update $\theta^{(t+1)}$ using the loss weighted by $\{\omega_i^{(t)}\}_{i=1}^n$ (Eq. \ref{eq:classifier_update})
    \State Update $\phi^{(t+1)}$ (Eq. \ref{eq:weight_network_update})
\EndFor

\end{algorithmic}
\begin{flushleft}
\textbf{Output}: Optimized weight network parameter $\phi^T$ 
\end{flushleft}
\vspace{-0.3em}
\end{algorithm}

\subsection{Model-Level Generation Capability Optimization (MLCO)}
\label{sec:mlpo}

Although standard diffusion models demonstrate remarkable capabilities in generating visually high-quality images, their outputs often fail to meet the specific data requirements of downstream applications. To address this misalignment, we propose an iterative DPO framework that adapts the generative model to downstream task-specific data preferences.  





Each optimization cycle begins by prompting the diffusion model with prompts $c_i$ of the form ``a photo of $[I_i]$'', where class-specific identifiers $[I_i]$ are obtained via textual inversion~\cite{gal2022image}, to generate a synthetic dataset $\mathcal{D}_{\text{syn}} = \{(x_i)\}_{i=1}^M$. The pre-trained weight network $\mathcal{W}_\phi$ subsequently evaluates each sample's utility via the weight score $\omega_i = \mathcal{W}_\phi(\mathcal{L}(f(x_i;\theta), y_i))$, where $\mathcal{L}(f(x_i;\theta), y_i)$ is the loss of the classifier $f_\theta$ on sample $x_i$ with its label $y_i$. Based on these scores, high-utility samples $x_i^w$ and low-utility samples $x_i^l$ are paired to construct a preference dataset:

\begin{equation}
\mathcal{D}_{\text{preference}} = \{(c_i, x_i^w, x_i^l) \mid \mathcal{W}_\phi ( \mathcal{L}(f(x_i^w;\theta), y_i^w)) > \mathcal{W}_\phi ( \mathcal{L}(f(x_i^l;\theta), y_i^l)) \}_{i=1}^N ,
\end{equation}

We then use the preference dataset \( \mathcal{D}_{\text{preference}} \) to fine-tune the diffusion model's U-Net \( \psi \) using DPO, with the optimization objective formulated according to the Diffusion DPO ~\cite{wallace2024diffusion}. 
\begin{equation}
\begin{aligned}
\mathcal{L}_{\text{DPO}}(\psi) = & -\mathbb{E}_{\substack{(x_0^w, x_0^l) \sim \mathcal{D_{\text{preference}}},  t \sim \mathcal{U}(0,T),  x_t^w \sim q(x_t^w|x_0^w),  x_t^l \sim q(x_t^l|x_0^l)}} \\ 
& \left[ \log \sigma\left( -\beta T \omega(\lambda_t) \left( \Delta \mathcal{L}_w - \Delta \mathcal{L}_l \right) \right) \right],
\end{aligned}
\label{eq:dpo_loss}
\end{equation}
\begin{equation}
    \begin{aligned}
    \Delta \mathcal{L}_w &= \|\epsilon^w - \epsilon_\psi(x_t^w, t)\|^2 - \|\epsilon^w - \epsilon_{\text{ref}}(x_t^w, t)\|^2 \\
    \Delta \mathcal{L}_l &= \|\epsilon^l - \epsilon_\psi(x_t^l, t)\|^2 - \|\epsilon^l - \epsilon_{\text{ref}}(x_t^l, t)\|^2.
    \end{aligned} 
\end{equation}
Here, \(\epsilon_\psi\) and \(\epsilon_{\text{ref}}\) denote the noise predictions from the trainable and reference U-Nets, respectively.  The forward diffusion process \(q(x_t | x_0)\) adds noise to \(x_0\) at timestep \(t\), where \(t\) is sampled from \(\mathcal{U}(0, T)\). \(\beta\) balances preference alignment and KL regularization, while \(\sigma(\cdot)\) is the sigmoid activation in the loss. \(\lambda_t\) is the signal-to-noise ratio \cite{kingma2021variational} and \(\omega(\lambda_t)\) is weighting function \cite{ho2020denoising}.

Through iterative DPO fine-tuning, we progressively adapt the diffusion model's generative capability according to the downstream task's preferences for the training data. This process enables better alignment between the model's output distribution and the target application requirements.

\subsection{Instance-Level Generation Policy Optimization (ILPO)}

\label{sec:ilo}


While MLPO tailors the generative model to the downstream task at a coarse level, ILPO performs fine-grained refinement of the generation policy by jointly optimizing the prompt embeddings and the initial noise. The overall optimization objective is formulated as:
\begin{equation}
(p^*, \epsilon_T^*) = \arg\max_{p,\epsilon_T} \mathbb{E}\left[\mathcal{W}_\phi(\mathcal{L}(f(g(p,\epsilon_T);\theta), y))\right],
\end{equation}
where $p$ denotes the prompt embedding, $\epsilon_T$ represents the initial noise vector, $g(\cdot)$ is the diffusion model, $f(\cdot;\theta)$ is the downstream classifier, $y$ is the ground-truth label, and $\mathcal{L}(\cdot)$ is the classification loss. The optimization process consists of two synergistic components:

\textbf{Prompt embedding optimization:} Building upon the class-specific identifiers \( [I_i] \) learned through textual inversion \cite{gal2022image}, we optimize the prompt embeddings \( p_i = E(\text{``a photo of } [I_i] \text{'')}\) using gradient-based optimization, where \( E(\cdot) \) denotes the text encoder. This process aims to maximize the utility score predicted by the pre-trained weight network. To preserve semantic alignment during optimization, a CLIP-based regularization term 
$L_{\text{CLIP}} = -\cos(E(x_i), e_i)$ is applied, 
where $x_i$ is the generated image, 
$E(\cdot)$ is the CLIP image encoder and 
$e_i = \text{avg}(E(\{x_j\})$ is the mean embedding of a set of target-class real images $\{x_j\}$. The complete prompt optimization objective is:
\begin{equation}
    p^* = \arg\max_p \left[ \mathcal{W}_\phi(\mathcal{L}(f(g(p,\epsilon_T);\theta), y)) - \lambda L_{\text{CLIP}} \right],
    \label{eq:8}
\end{equation}

\vspace{-0.7em}
where $\lambda$ controls the regularization strength. This joint optimization enhances sample utility while preserving semantic coherence with the original class concept.
\begin{figure*}[t] 
    \centering
    \includegraphics[width=\textwidth]{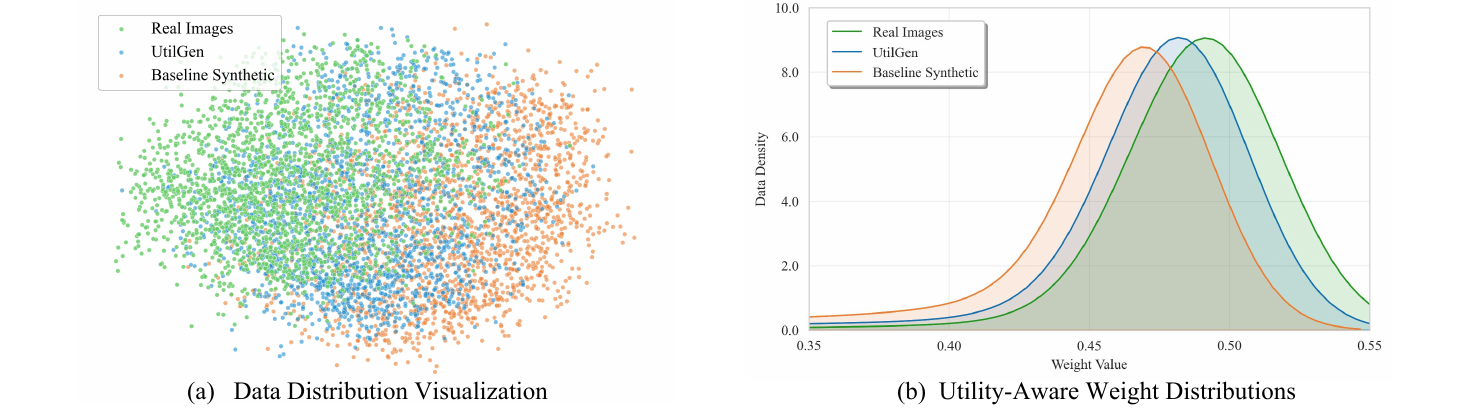}
    \caption{  (a) Feature space visualization on the Flower dataset ~\cite{nilsback2008automated}shows that our synthetic data achieves closer alignment with the real data distribution compared to vanilla Stable Diffusion.
        (b) Utility-aware weight distributions for synthetic and real data on the Flower dataset~\cite{nilsback2008automated}, showing sample utility scores for downstream tasks.}
    \label{fig:distribution} 
\end{figure*}

\begin{algorithm}[t] 

\caption{Instance-Level Generation Policy Optimization (ILPO)}\label{algo_ilo}
\footnotesize

\begin{flushleft}
\vspace{-0.3em}
\textbf{Input}: $K$ classes: $\{c_1, ..., c_K\}$;  number of images to generate for each class: $\{N_1, \dots, N_K\}$, where $N_k$ is the number of samples to generate for class $c_k$

\textbf{Required}: Diffusion model $g(\cdot,\cdot)$ fine-tuned by MLCO; Textual Inversion technology $TI(\cdot,\cdot)$ using few-shot real images per class $\mathcal{D}_k = \{x_1^k, ..., x_N^k\}$ for each $c_k$;  
\end{flushleft}

\begin{algorithmic}[1]

\For{each class $k \in \{1,...,K\}$}
    \State \textbf{// Prompt Embedding Optimization}
    \State Initialize $p_k \gets TI(c_k, \mathcal{D}_k)$
    \State $p_k^* \gets p_k$ (Eq.~\ref{eq:8}) \Comment{Optimize to maximize data utility}

    \State \textbf{// Noise Optimization}
    \For{$i = 1$ to $N_k$}
        \State Sample noise $\epsilon_T \sim \mathcal{N}(0,I)$
        \State $\epsilon'_T \gets \epsilon_T$ by (Eq.~\ref{eq:9}) \Comment{Inject semantic information representing high-utility data}
        \State $x'_i \gets g(p_k^*, \epsilon'_T)$ \Comment{Generate high-utility data}
        \State $\mathcal{D}_{\text{synth}} \gets \mathcal{D}_{\text{synth}} \cup \{x'_i\}$ 
    \EndFor
\EndFor

\end{algorithmic}

\begin{flushleft}
\textbf{Output}: High-utility synthetic dataset $\mathcal{D}_{\text{synth}}$
\end{flushleft}

\end{algorithm}
\textbf{Noise optimization:} 
The quality of synthetic images is influenced by both the text prompt and the random Gaussian noise. Yet, since each image generation requires independently sampled noise, directly optimizing noise vectors via gradient ascent to improve utility is computationally prohibitive. Recent studies~\cite{zhou2024golden, bai2024zigzag, ahn2024noise} show that the discrepancy between denoising and inversion Classifier-Free Guidance (CFG) scales can be leveraged to implicitly inject prompt semantic information into the initial noise. Leveraging this observation, we adapt the methodology to optimize the initial noise, enabling it to incorporate semantic information of high-utility data. Formally, the noise optimization process is defined as:
\begin{equation}
\epsilon'_t = \text{DDIM-Inversion}_{\omega_w}(\text{DDIM}_{\omega_l}(\epsilon_t, p^*)).
\label{eq:9}
\end{equation}
where $\omega_l$ and $\omega_w$ are the CFG scales for the denoising process $\text{DDIM}(\cdot)$ and the inversion process $\text{DDIM-Inversion}(\cdot)$, respectively. The condition $\omega_l > \omega_w$ enables the implicit injection of semantic information into the initial noise. The entire process of ILPO is outlined in Algorithm \ref{algo_ilo}.

The synergistic integration of MLCO (Sec.\ref{sec:mlpo}) and ILPO (Sec.\ref{sec:ilo}) enables \textsc{UtilGen} to synthesize data that closely aligns with real data feature distributions, as visualized in Fig.\ref{fig:distribution}(a). This approach simultaneously achieves higher utility scores (Fig.\ref{fig:distribution}(b)), demonstrating enhanced task relevance and superior data utility in downstream applications.


\section{Experiments}

\begin{table}[t]
  \centering
  \footnotesize
  \scriptsize
  \setlength{\tabcolsep}{2mm}
  \renewcommand{\arraystretch}{1.1}
  \caption{Classification performance across eight datasets using ResNet-50 \cite{he2016deep} as the backbone classifier. \textsc{UtilGen} and DataDream \cite{kim2024datadream} generate synthetic data guided by \textbf{16-shot real images} per class, while other methods use the full real dataset as guidance for generation. The top block shows results using synthetic data only, while the bottom includes joint training (synthetic + full real dataset). Each method produces synthetic data at 5$\times$ the scale of the original real dataset. }
  \label{tab:performance}
  \begin{tabular}{l|cc|cccccccc|c}
    \toprule
    Method & Real & Syn & IN-1k-S & IN-100-S & Cal101 & DTD & CUB & PETs & Food-S & Flowers & Avg \\
    \midrule

     \rowcolor{gray!40} \multicolumn{12}{c}{Training on Synthetic Data Only} \\ \midrule
    SD v2.1 \cite{rombach2022high}   &            & \checkmark & 24.35 & 27.96 & 14.74 &  7.92 & 23.43 & 26.05 & 24.02 & 31.41 & 22.49 \\
    GIF \cite{zhang2023expanding}      &            & \checkmark & 28.95 & 31.94 & 20.39 & 13.19 & 27.54 & 29.73 & 25.94 & 56.12 & 29.23 \\
    GAP \cite{yeo2024controlled}       &            & \checkmark & 25.84 & 30.94 & 18.70 & 11.01 & 29.49  & 27.46 & 25.31 & 53.66 & 27.80 \\
    DataDream \cite{kim2024datadream} &            & \checkmark & 30.35 & 35.48 & 23.61 & 13.24 & 35.38 & 34.77 & 28.41 & 65.15 & 33.30 \\
    \textbf{UtilGen}
               &            & \textbf{\checkmark} & \textbf{33.72} & \textbf{40.94} & \textbf{29.31} & \textbf{13.52} & \textbf{43.32} & \textbf{37.25} & \textbf{31.87} & \textbf{67.43} & \textbf{37.17} \\

   \multicolumn{3}{l|}{\textbf{\textit{ $\triangle$ over previous SOTA}}} & \textcolor{green!70!black}{\textbf{+3.37}} & \textcolor{green!70!black}{\textbf{+5.46}} & \textcolor{green!70!black}{\textbf{+5.70}} & \textcolor{green!70!black}{\textbf{+0.28}} & \textcolor{green!70!black}{\textbf{+7.94}} & \textcolor{green!70!black}{\textbf{+2.48}} & \textcolor{green!70!black}{\textbf{+3.46}} & \textcolor{green!70!black}{\textbf{+2.28}} & \textcolor{green!70!black}{\textbf{+3.87}} \\

    \midrule

    \rowcolor{gray!40} \multicolumn{12}{c}{Joint Training with Real Data} \\ \midrule
    Real Dataset      & \checkmark &            & 36.34 & 38.58 & 43.55 & 16.32 & 21.03 & 28.78 & 19.88 & 73.34 & 34.73 \\
    SD v2.1 \cite{rombach2022high}      & \checkmark & \checkmark & 43.26 & 49.86 & 59.35 & 28.82 & 43.52 & 52.35 & 40.32 & 79.58 & 49.63 \\
    GIF \cite{zhang2023expanding}          & \checkmark & \checkmark & 49.85 & 54.12 & 67.60 & 33.45 & 43.80 & 58.21 & 41.39 & 84.47 & 54.11 \\
    GAP \cite{yeo2024controlled}          & \checkmark & \checkmark & 46.58 & 53.14 & 66.97 & 32.87 & 47.46 & 57.86 & 43.77 & 84.84 & 54.19 \\
    DataDream \cite{kim2024datadream}    & \checkmark & \checkmark & 52.16 & 57.68 & 73.38 & 34.84 & 53.43 & 60.83 & 47.44 & 89.60 & 58.67 \\
    \textbf{UtilGen}
                  & \textbf{\checkmark} & \textbf{\checkmark} & \textbf{54.56} & \textbf{61.54} & \textbf{75.62} & \textbf{36.06} & \textbf{57.53} & \textbf{64.64} & \textbf{52.72} & \textbf{93.62} & \textbf{62.04} \\

    \multicolumn{3}{l|}{\textbf{\textit{ $\triangle$ over real dataset}}} & \textcolor{green!70!black}{\textbf{+18.22}} & \textcolor{green!70!black}{\textbf{+22.96}} & \textcolor{green!70!black}{\textbf{+32.07}} & \textcolor{green!70!black}{\textbf{+19.74}} & \textcolor{green!70!black}{\textbf{+36.50}} & \textcolor{green!70!black}{\textbf{+35.86}} & \textcolor{green!70!black}{\textbf{+32.84}} & \textcolor{green!70!black}{\textbf{+20.28}} & \textcolor{green!70!black}{\textbf{+27.31}} \\
\multicolumn{3}{l|}{\textbf{\textit{ $\triangle$ over previous SOTA}}} & \textcolor{green!70!black}{\textbf{+2.40}} & \textcolor{green!70!black}{\textbf{+3.86}} & \textcolor{green!70!black}{\textbf{+2.24}} & \textcolor{green!70!black}{\textbf{+1.22}} & \textcolor{green!70!black}{\textbf{+4.10}} & \textcolor{green!70!black}{\textbf{+3.81}} & \textcolor{green!70!black}{\textbf{+5.28}} & \textcolor{green!70!black}{\textbf{+4.02}} & \textcolor{green!70!black}{\textbf{+3.37}} \\
   
    \bottomrule
  \end{tabular}
  \vspace{-2em}
  
\end{table}

\subsection{Experimental Setup}

\textbf{Benchmarks.}  We evaluate the effectiveness of \textsc{UtilGen} across eight datasets spanning three classification tasks: coarse-grained classification (ImageNet-1k-Subset~\cite{deng2009imagenet}, ImageNet-100-Subset~\cite{deng2009imagenet}, and Caltech 101~\cite{fei2004learning}), fine-grained classification (Oxford Pets~\cite{parkhi2012cats}, Food-S~\cite{bossard2014food}, Flowers 102~\cite{nilsback2008automated}, and CUB-200-2011~\cite{wah2011caltech}), and texture classification (DTD~\cite{cimpoi2014describing}). 
Specifically, ImageNet-1k-Subset and Food-S are subsets of ImageNet-1K and Food101~\cite{bossard2014food}, respectively, each containing 100 randomly selected images per class. ImageNet-100-Subset is constructed by randomly sampling 100 animal-related classes from the original ImageNet-1K~\cite{deng2009imagenet}, with 100 randomly selected images per class. 
Further benchmark details are provided in Appendix~\ref{tab:dataset_detail}.

\textbf{Baselines.} To compare our utility-centric approach with existing methods that focus on optimizing intrinsic data characteristics, we select GIF \cite{zhang2023expanding} and DataDream \cite{kim2024datadream} as representative baselines. Specifically, GIF \cite{zhang2023expanding} enhances diversity by applying feature perturbations, while DataDream \cite{kim2024datadream} improves fidelity through domain alignment using LoRA fine-tuning of the diffusion model. Additionally, we include GAP \cite{yeo2024controlled}, which uses feedback from the downstream model to generate adversarial prompts that maximize the model's loss on the generated images. This enables a direct comparison between its loss-based feedback strategy and the utility-based feedback mechanism employed by \textsc{UtilGen}. For fair comparison, we adopt Stable Diffusion v2.1 \cite{rombach2022high} (SD v2.1) as the backbone generator across all baseline methods. The implementation details of \textsc{UtilGen} are in Appendix~\ref{imp_detail}.

\subsection{Evaluation Results}
\label{sec:comparison}

\noindent \textbf{Results on solely synthetic data.} As shown in Table~\ref{tab:performance} (top), \textsc{UtilGen} achieves the highest average accuracy of 37.17\% when trained solely on synthetic data, outperforming the previous best method DataDream (33.30\%) by a notable margin of +3.87\%. It demonstrates strong performance across both coarse-grained (e.g., 40.94\% on IN-100-S~\cite{deng2009imagenet}) and fine-grained tasks (e.g., 67.43\% on Flowers~\cite{nilsback2008automated}), indicating excellent generalization despite training solely on synthetic data.

\noindent \textbf{Results on real + synthetic Data.}  
In the joint training setting (bottom of Table~\ref{tab:performance}), \textsc{UtilGen} maintains its lead with an average accuracy of 62.04\%, surpassing DataDream (58.67\%) by +3.54\%. It achieves particularly strong gains on fine-grained datasets (e.g., 93.62\% on Flowers~\cite{nilsback2008automated}), while also delivering consistent improvements on coarse-grained tasks. These results suggest that \textsc{UtilGen} can effectively complement real data across different task granularities, providing high-utility synthetic samples that enhance model performance and robustness.

\begin{figure*}[t] 
    \centering
    \includegraphics[width=\textwidth]{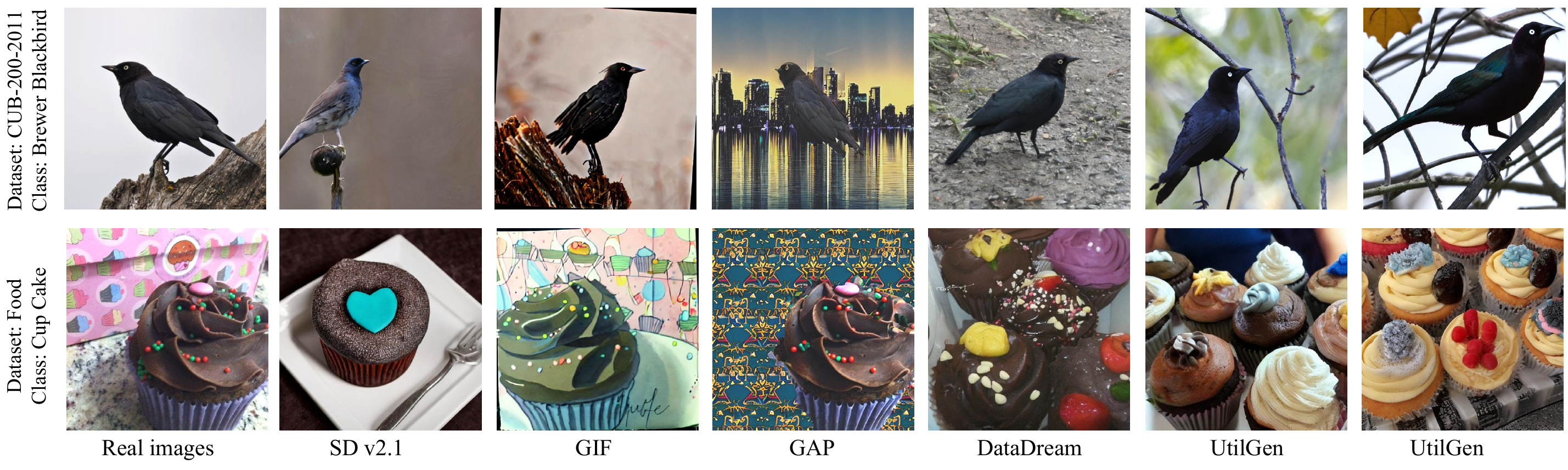}
    \caption{Comparison of synthetic images generated by SD v2.1, GIF \cite{zhang2023expanding}, GAP \cite{yeo2024controlled}, DataDream \cite{kim2024datadream}, and \textsc{UtilGen}. }
    \label{fig:visual} 
\end{figure*}

\begin{table*}[t]
\centering
\setlength{\tabcolsep}{4.5pt}
\renewcommand{\arraystretch}{1.2}

\begin{minipage}[t]{0.565\textwidth}
\centering
\caption{Performance comparison under different synthesis budgets on ImageNet-100. \textsc{UtilGen} scales more effectively with increasing synthetic data ratios.}
\label{tab:scaling_comparison}
\scriptsize
\begin{tabular}{lccccc}
\toprule
Budget & SD v2.1 & GIF~\cite{zhang2023expanding} & GAP~\cite{yeo2024controlled} & DataDream~\cite{kim2024datadream} & \textsc{UtilGen} \\
\midrule
1$\times$ & 12.18 & 13.44 & 15.02 & 17.12 & \textbf{18.04} \\
3$\times$ & 20.20 & 21.90 & 21.16 & 25.26 & \textbf{28.52} \\
5$\times$ & 27.96 & 31.94 & 30.94 & 35.48 & \textbf{40.94} \\
\bottomrule
\end{tabular}
\end{minipage}
\hfill
\begin{minipage}[t]{0.415\textwidth}
\centering
\caption{Comparison of intra-class diversity of synthetic data on ImageNet-100. Higher values indicate greater diversity.}
\label{tab:diversity}
\scriptsize
\begin{tabular}{lc}
\toprule
Method & Mean Intra-Class Diversity $\uparrow$ \\
\midrule
Stable Diffusion v2.1 & 0.5815 \\
DataDream~\cite{kim2024datadream} & 0.5238 \\
\textsc{UtilGen} & \textbf{0.6054} \\
\bottomrule
\end{tabular}
\end{minipage}
\end{table*}

\begin{figure}[t]
    \centering
    \begin{minipage}[t]{0.48\linewidth}
        \centering
        \includegraphics[width=\linewidth,height=3cm]{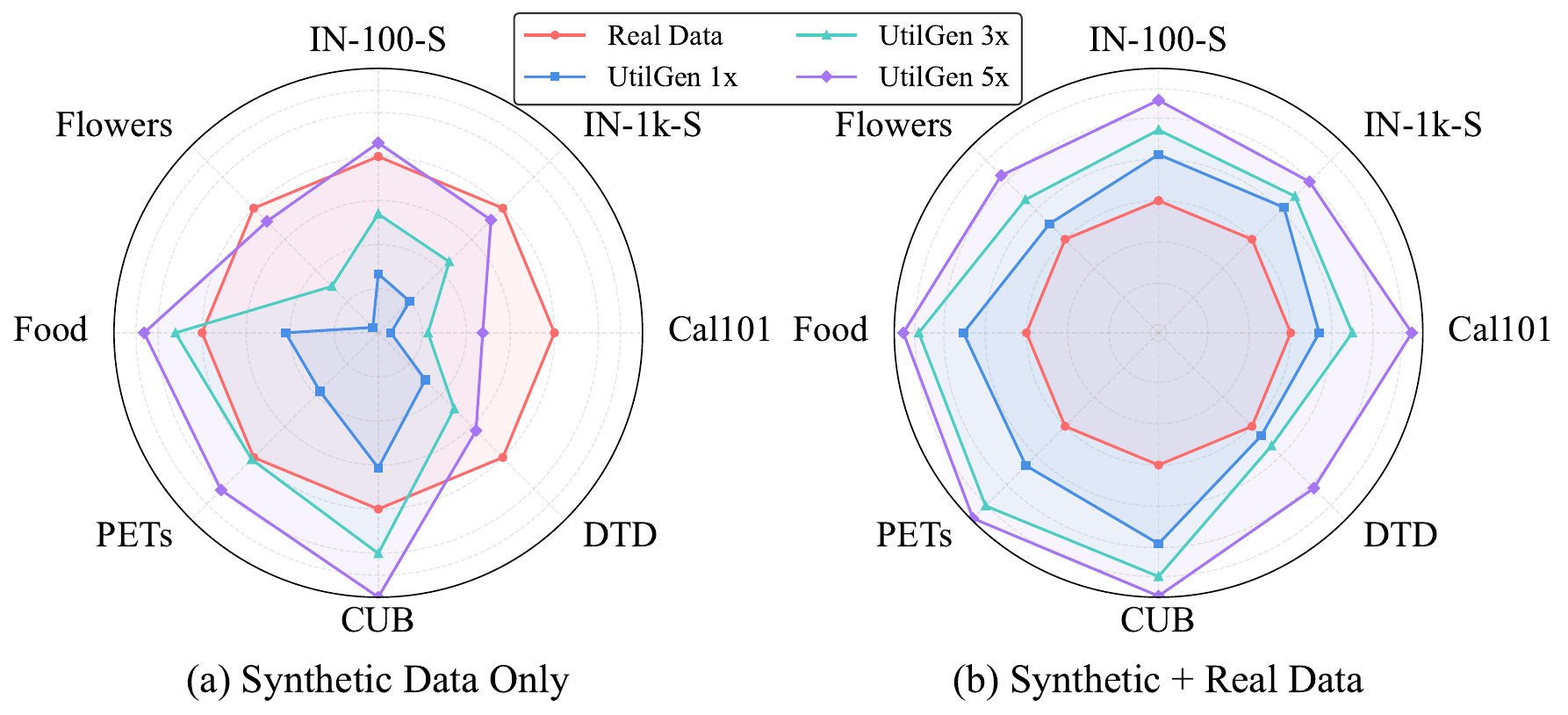}
        \caption{Synthetic Data Scaling Effects across different training data regimes. (a) Models trained exclusively on synthetic data (Synthetic Data Only). (b) Models trained on combined synthetic and real data (Synthetic + Real Data).}
        \label{fig:scaling_effect}
    \end{minipage}
    \hfill
    \begin{minipage}[t]{0.48\linewidth}
        \centering
        \centering
        \includegraphics[width=\linewidth,height=3cm]{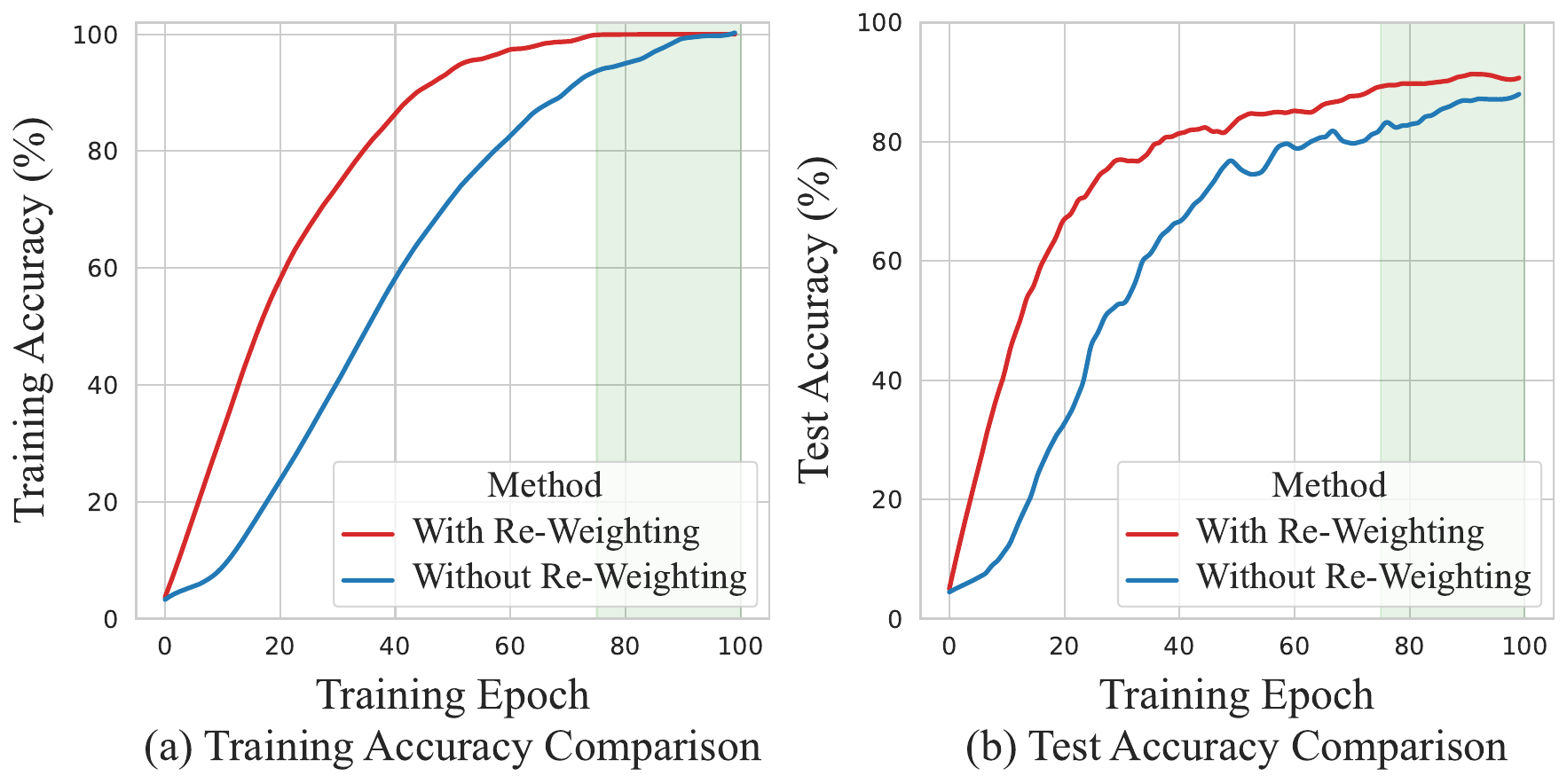}
        \caption{Training and test accuracy comparison with and without the weight network using ResNet-50 \cite{he2016deep} on the dataset (original Flowers + SD v2.1 augmented data with 1$\times$ expansion). (a) Training convergence; (b) Test accuracy.}
        \label{fig:accuracy_plot}
    \end{minipage}
    \vspace{-0.3em}
\end{figure}


\noindent \textbf{Synthetic data scaling effects.}  
\textsc{UtilGen} exhibits strong scalability and data augmentation efficiency. As shown in Figure~\ref{fig:scaling_effect}, scaling synthetic data from 1$\times$ to 5$\times$ the original set consistently improves ResNet-50~\cite{he2016deep} performance across benchmarks under both synthetic-only and joint training settings. When trained with 3$\times$ synthetic data alone, models outperform their real-data-trained counterparts on three datasets; at 5$\times$ scaling, this advantage extends to four datasets. To further examine how performance scales with the synthesis budget, we compare \textsc{UtilGen} with four representative baselines under budgets. As shown in Table~\ref{tab:scaling_comparison}, \textsc{UtilGen} consistently achieves the best results across all budgets, and the performance gap widens as the synthesis ratio increases, demonstrating superior scalability under large-scale generation. Additional details on augmentation efficiency and computational cost are provided in Appendix~\ref{appendix:scaling}.

\noindent \textbf{Diversity analysis of synthetic data.}  
We evaluate the diversity of synthetic samples by computing the mean intra-class cosine distance of CLIP (ViT-L/14) features on ImageNet-100. Compared methods include vanilla Stable Diffusion v2.1, DataDream~\cite{kim2024datadream}, and \textsc{UtilGen}. Higher values indicate greater intra-class diversity, which typically benefits generalization.  As shown in Table~\ref{tab:diversity}, \textsc{UtilGen} achieves the highest intra-class diversity, indicating that its utility-guided optimization preserves sample variety and avoids mode collapse.


\begin{table*}[!t]
\centering
\setlength{\tabcolsep}{2.5pt}
\renewcommand{\arraystretch}{1.15}

\begin{minipage}[t]{0.51\textwidth}
\centering
\caption{Reusability of synthetic data across downstream models on ImageNet-100.}
\label{tab:cross_task}
\scriptsize
\begin{tabular}{lccc}
\toprule
\makecell[l]{Classifier used to \\ train weight network} &
\makecell[c]{Downstream model \\ using synthetic data} &
Method & Acc. (\%) \\
\midrule
-- & WideResNet & DataDream~\cite{kim2024datadream} & 31.76 \\
ResNet-50 & WideResNet & \textsc{UtilGen} & \textbf{36.40} \\
-- & CLIP & DataDream~\cite{kim2024datadream} & 71.42 \\
ResNet-50 & CLIP & \textsc{UtilGen} & \textbf{72.14} \\
\bottomrule
\end{tabular}
\end{minipage}
\hfill
\begin{minipage}[t]{0.48\textwidth}
\centering
\caption{Generalization performance across diverse network architectures on ImageNet-100.}
\label{tab:arch_comparison}
\scriptsize
\begin{tabular}{lccc}
\toprule

\multirow{2}{*}{Method} & \multirow{2}{*}{ResNeXt-50} & \multirow{2}{*}{WideResNet-50} & \multirow{2}{*}{MobileNetV2} \\[2ex]

\midrule

GIF~\cite{zhang2023expanding} & 27.54 & 27.84 & 31.24 \\
GAP~\cite{yeo2024controlled} & 27.66 & 27.76 & 32.72 \\
DataDream~\cite{kim2024datadream} & 31.24 & 31.76 & 35.48 \\
\textbf{\textsc{UtilGen}} & \textbf{37.62} & \textbf{37.82} & \textbf{40.59} \\
\bottomrule
\end{tabular}
\end{minipage}
\end{table*}

\noindent \textbf{Reusability of synthetic data across tasks.}  
We further analyze the reusability of \textsc{UtilGen}-generated data across different downstream models on the ImageNet-100 dataset. Even when the weight network is trained using ResNet-50 , the resulting synthetic data generalizes effectively to other models such as WideResNet and CLIP. As shown in Table~\ref{tab:cross_task}, the performance gains remain consistent across architectures, indicating that high-utility samples identified by \textsc{UtilGen} are not tied to a specific model and can be reused for diverse learning objectives.

\noindent \textbf{Generalization across different architectures.}  
To assess the versatility of \textsc{UtilGen}, we evaluate its performance on three architectures: ResNeXt-50 \cite{xie2017aggregated}, WideResNet-50 \cite{zagoruyko2016wide}, and MobileNetV2 \cite{sandler2018mobilenetv2}. As shown in Table~\ref{tab:arch_comparison}, \textsc{UtilGen} consistently achieves the highest accuracy, with 500 images generated per class, confirming the effectiveness of our approach across various network architectures.

\noindent \textbf{Cost-benefit analysis compared to collecting more labeled data.}  
Compared to manual data collection and annotation, \textsc{UtilGen} offers a highly cost-effective solution for dataset expansion. According to the Masterpiece Group\footnote{\url{https://mpg-myanmar.com/annotation/}}, manually annotating 10,000 images (e.g., 100 images per class across 100 classes) typically takes about two weeks and costs approximately \$800. In contrast, generating the same amount of data using \textsc{UtilGen} requires only about 0.94 hours and \$20 on 8 V100 GPUs rented from Google Cloud\footnote{\url{https://cloud.google.com/compute/gpus-pricing}}. On the ImageNet-100-Subset dataset, using 5$\times$ synthetic data produced by \textsc{UtilGen} even surpasses the real-data baseline in accuracy, while requiring only $\sim$4.7 hours and about \$100 in compute cost, as shown in Table~\ref{tab:cost_comparison}.

\noindent \textbf{Ablation study. } 
Table~\ref{tab:ablation} presents an ablation study on the IN-100-S dataset using ResNet-50 trained solely on synthetic data (500  aasssper class). We analyze the effects of MLCO and ILPO (including prompt embedding and initial noise optimization). Each component brings improvements over the baseline, and combining all three achieves the best performance, surpassing the baseline by +12.98\%, demonstrating the complementary strengths of model-level and instance-level optimizations.

\subsection{Mechanism Analysis}

\textbf{Effect of sample re-weighting on classifier training. } To assess the effect of the weight allocation network on classifier training, we compare the training trajectories of models with and without dynamic weighting. As shown in Figure~\ref{fig:accuracy_plot}(a), the weighted model converges faster and achieves higher training accuracy in earlier epochs, suggesting it effectively assigns higher weights to high-utility samples during learning. Figure~\ref{fig:accuracy_plot}(b) presents the test accuracy curves, where accuracy rises more rapidly and achieves a final improvement of 2.94\% over the baseline. This gain is attributed to the network’s ability to down-weight low-utility or noisy samples, thereby mitigating their negative impact. These results confirm that the pre-trained weight allocation network can reliably measure sample utility and serve as a signal to guide synthetic data generation.

\begin{figure*}[t] 
    \centering
    \includegraphics[width=\textwidth]{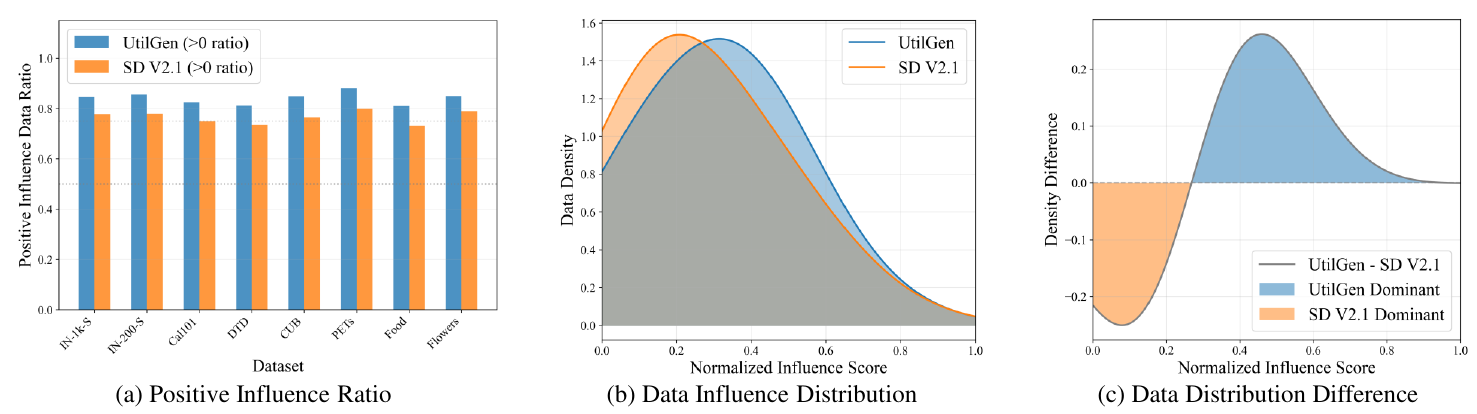}
    \caption{Data influence comparison between \textsc{UtilGen} and SD v2.1, computed using influence functions.  
(a) Proportion of samples with positive influence scores (influence > 0) across eight datasets. 
(b) Influence score distributions showing that \textsc{UtilGen} generates a higher density of samples with stronger influence. 
(c) Data density difference plot highlighting that \textsc{UtilGen} dominates in high-influence regions, while SD v2.1 contributes more to low-influence areas.}
    \vspace{-1.0em}
    \label{fig:inf} 
\end{figure*}

\noindent \textbf{Data influence analysis.}
We measure data influence by jointly training models on synthetic datasets generated by \textsc{UtilGen} and the baseline SD v2.1, applying Influence Function~\cite{koh2017understanding}. Figure~\ref{fig:inf}(a) reveals that \textsc{UtilGen} consistently produces a higher proportion of positively influential samples (influence > 0) across all benchmarks. Furthermore, the influence score distribution shown in Figure~\ref{fig:inf}(b) shifts noticeably to the right, indicating that \textsc{UtilGen} generates more samples with stronger positive influence. Complementing this, Figure~\ref{fig:inf}(c) highlights that \textsc{UtilGen} achieves substantially higher data density within high-influence regions, while simultaneously reducing sample concentration in low-utility areas. Together, these results validate the effectiveness of our method in synthesizing impactful data that better supports downstream model optimization.

\begin{table}[t]
\centering
\scriptsize
\begin{minipage}{0.52\textwidth}
\centering
\fontsize{8}{10.5}\selectfont
\caption{Comparison of the manual annotation paradigm and our synthetic data paradigm in terms of cost, time, and performance.}
\vspace{0.4em}
\label{tab:cost_comparison}
\setlength{\tabcolsep}{2pt}
\renewcommand{\arraystretch}{1.7}
\begin{tabular}{lcccc}
\Xhline{0.8pt}
Method & Images & Time Required & Cost &  Acc. (\%) \\
\hline
Manual Annotation & 10,000 & $\sim$2 weeks & \$800 & 38.58 \\
\textsc{UtilGen} (1$\times$) & 10,000 & $\sim$0.94 h & \$20 & 18.04 \\
\textsc{UtilGen} (5$\times$) & 50,000 & $\sim$4.70 h & \$100 & \textbf{40.94} \\
\Xhline{0.8pt}
\end{tabular}
\end{minipage}
\hfill
\begin{minipage}{0.46\textwidth}
\centering
\caption{Ablation study on ImageNet-100.}
\vspace{0.2em}
\label{tab:ablation}
\setlength{\tabcolsep}{3pt}
\renewcommand{\arraystretch}{1.3}
\begin{tabular}{lcc|c}
\Xhline{0.8pt}
MLCO & Prompt Optimization & Noise Optimization & Acc. (\%) \\
\hline
 & & &  27.96 \\
\hline
& & \checkmark &  28.68\\
\checkmark & & & 36.42 \\
 & \checkmark & & 37.96 \\
\hline
\checkmark & & \checkmark & 32.08 \\
 & \checkmark & \checkmark & 39.12 \\
\checkmark & \checkmark & & 39.73 \\
\hline
\checkmark & \checkmark & \checkmark & \textbf{40.94} \\
\Xhline{0.8pt}
\end{tabular}
\end{minipage}
\vspace{-0.4em}
\end{table}

\section{Conclusion}

In this study, we propose \textsc{UtilGen}, a utility-centric data augmentation framework that shifts the focus from optimizing intrinsic visual properties to enhancing task-specific utility. By incorporating downstream model feedback, \textsc{UtilGen} adaptively adjusts the data generation process to produce high-utility data tailored for specific downstream tasks, thereby establishing a feedback loop between data generation and model training. Experiments across eight benchmarks demonstrate consistent performance gains,  and in certain cases surpassing the performance of models trained solely on real data. These results highlight the superiority of the utility-centric approach over prior methods focusing primarily on intrinsic visual quality. These findings underscore the potential of utility-centric generation and suggest that integrating task-specific utility alongside traditional visual quality considerations offers a more effective paradigm for future data augmentation research.

\section*{Acknowledgment}
The authors would like to acknowledge the support from the following funding sources. Shuo Yang is supported by the National Natural Science Foundation of China Young Scientists Fund (No. 62506096). Xiaobo Xia is partially supported by the MoE Key Laboratory of Brain-inspired Intelligent Perception and Cognition at the University of Science and Technology of China (Grant No. 2421002). Bo Zhao is supported by the National Natural Science Foundation of China (Grant No. 62306046).

\bibliographystyle{unsrt}
\bibliography{custom}

\begin{thebibliography}{10}

\bibitem{ho2020denoising}
Jonathan Ho, Ajay Jain, and Pieter Abbeel.
\newblock Denoising diffusion probabilistic models.
\newblock {\em Advances in neural information processing systems}, 33:6840--6851, 2020.

\bibitem{dhariwal2021diffusion}
Prafulla Dhariwal and Alexander Nichol.
\newblock Diffusion models beat gans on image synthesis.
\newblock {\em Advances in neural information processing systems}, 34:8780--8794, 2021.

\bibitem{wang2025lavin}
Zhaoqing Wang, Xiaobo Xia, Runnan Chen, Dongdong Yu, Changhu Wang, Mingming Gong, and Tongliang Liu.
\newblock Lavin-dit: Large vision diffusion transformer.
\newblock In {\em Proceedings of the Computer Vision and Pattern Recognition Conference}, pages 20060--20070, 2025.

\bibitem{rombach2022high}
Robin Rombach, Andreas Blattmann, Dominik Lorenz, Patrick Esser, and Bj{\"o}rn Ommer.
\newblock High-resolution image synthesis with latent diffusion models.
\newblock In {\em Proceedings of the IEEE/CVF conference on computer vision and pattern recognition}, pages 10684--10695, 2022.

\bibitem{ho2022classifier}
Jonathan Ho and Tim Salimans.
\newblock Classifier-free diffusion guidance.
\newblock {\em arXiv preprint arXiv:2207.12598}, 2022.

\bibitem{luo2024deem}
Run Luo, Yunshui Li, Longze Chen, Wanwei He, Ting-En Lin, Ziqiang Liu, Lei Zhang, Zikai Song, Xiaobo Xia, Tongliang Liu, et~al.
\newblock Deem: Diffusion models serve as the eyes of large language models for image perception.
\newblock {\em arXiv preprint arXiv:2405.15232}, 2024.

\bibitem{song2020denoising}
Jiaming Song, Chenlin Meng, and Stefano Ermon.
\newblock Denoising diffusion implicit models.
\newblock {\em arXiv preprint arXiv:2010.02502}, 2020.

\bibitem{he2022synthetic}
Ruifei He, Shuyang Sun, Xin Yu, Chuhui Xue, Wenqing Zhang, Philip Torr, Song Bai, and Xiaojuan Qi.
\newblock Is synthetic data from generative models ready for image recognition?
\newblock {\em arXiv preprint arXiv:2210.07574}, 2022.

\bibitem{hu2022lora}
Edward~J Hu, Yelong Shen, Phillip Wallis, Zeyuan Allen-Zhu, Yuanzhi Li, Shean Wang, Lu~Wang, Weizhu Chen, et~al.
\newblock Lora: Low-rank adaptation of large language models.
\newblock {\em ICLR}, 1(2):3, 2022.

\bibitem{yuan2023real}
Jianhao Yuan, Jie Zhang, Shuyang Sun, Philip Torr, and Bo~Zhao.
\newblock Real-fake: Effective training data synthesis through distribution matching.
\newblock {\em arXiv preprint arXiv:2310.10402}, 2023.

\bibitem{kim2024datadream}
Jae~Myung Kim, Jessica Bader, Stephan Alaniz, Cordelia Schmid, and Zeynep Akata.
\newblock Datadream: Few-shot guided dataset generation.
\newblock In {\em European Conference on Computer Vision}, pages 252--268. Springer, 2024.

\bibitem{zhang2023expanding}
Yifan Zhang, Daquan Zhou, Bryan Hooi, Kai Wang, and Jiashi Feng.
\newblock Expanding small-scale datasets with guided imagination.
\newblock {\em Advances in neural information processing systems}, 36:76558--76618, 2023.

\bibitem{da2023diversified}
Victor G~Turrisi da~Costa, Nicola Dall'Asen, Yiming Wang, Nicu Sebe, and Elisa Ricci.
\newblock Diversified in-domain synthesis with efficient fine-tuning for few-shot classification.
\newblock {\em arXiv preprint arXiv:2312.03046}, 2023.

\bibitem{zha2025data}
Daochen Zha, Zaid~Pervaiz Bhat, Kwei-Herng Lai, Fan Yang, Zhimeng Jiang, Shaochen Zhong, and Xia Hu.
\newblock Data-centric artificial intelligence: A survey.
\newblock {\em ACM Computing Surveys}, 57(5):1--42, 2025.

\bibitem{shu2019meta}
Jun Shu, Qi~Xie, Lixuan Yi, Qian Zhao, Sanping Zhou, Zongben Xu, and Deyu Meng.
\newblock Meta-weight-net: Learning an explicit mapping for sample weighting.
\newblock {\em Advances in neural information processing systems}, 32, 2019.

\bibitem{shu2023cmw}
Jun Shu, Xiang Yuan, Deyu Meng, and Zongben Xu.
\newblock Cmw-net: Learning a class-aware sample weighting mapping for robust deep learning.
\newblock {\em IEEE Transactions on Pattern Analysis and Machine Intelligence}, 45(10):11521--11539, 2023.

\bibitem{zhang2021learning}
Zizhao Zhang and Tomas Pfister.
\newblock Learning fast sample re-weighting without reward data.
\newblock In {\em Proceedings of the IEEE/CVF International Conference on Computer Vision}, pages 725--734, 2021.

\bibitem{he2016deep}
Kaiming He, Xiangyu Zhang, Shaoqing Ren, and Jian Sun.
\newblock Deep residual learning for image recognition.
\newblock In {\em Proceedings of the IEEE conference on computer vision and pattern recognition}, pages 770--778, 2016.

\bibitem{xie2017aggregated}
Saining Xie, Ross Girshick, Piotr Doll{\'a}r, Zhuowen Tu, and Kaiming He.
\newblock Aggregated residual transformations for deep neural networks.
\newblock In {\em Proceedings of the IEEE conference on computer vision and pattern recognition}, pages 1492--1500, 2017.

\bibitem{zagoruyko2016wide}
Sergey Zagoruyko and Nikos Komodakis.
\newblock Wide residual networks.
\newblock {\em arXiv preprint arXiv:1605.07146}, 2016.

\bibitem{sandler2018mobilenetv2}
Mark Sandler, Andrew Howard, Menglong Zhu, Andrey Zhmoginov, and Liang-Chieh Chen.
\newblock Mobilenetv2: Inverted residuals and linear bottlenecks.
\newblock In {\em Proceedings of the IEEE conference on computer vision and pattern recognition}, pages 4510--4520, 2018.

\bibitem{zhao2020dataset}
Bo~Zhao, Konda~Reddy Mopuri, and Hakan Bilen.
\newblock Dataset condensation with gradient matching.
\newblock {\em arXiv preprint arXiv:2006.05929}, 2020.

\bibitem{zhao2023dataset}
Bo~Zhao and Hakan Bilen.
\newblock Dataset condensation with distribution matching.
\newblock In {\em Proceedings of the IEEE/CVF Winter Conference on Applications of Computer Vision}, pages 6514--6523, 2023.

\bibitem{cao2023data}
Xiaofeng Cao, Weiyang Liu, and Ivor~W Tsang.
\newblock Data-efficient learning via minimizing hyperspherical energy.
\newblock {\em IEEE Transactions on Pattern Analysis and Machine Intelligence}, 45(11):13422--13437, 2023.

\bibitem{yang2024mind}
Shuo Yang, Zhe Cao, Sheng Guo, Ruiheng Zhang, Ping Luo, Shengping Zhang, and Liqiang Nie.
\newblock Mind the boundary: Coreset selection via reconstructing the decision boundary.
\newblock In {\em Forty-first International Conference on Machine Learning}, 2024.

\bibitem{yang2023bicro}
Shuo Yang, Zhaopan Xu, Kai Wang, Yang You, Hongxun Yao, Tongliang Liu, and Min Xu.
\newblock Bicro: Noisy correspondence rectification for multi-modality data via bi-directional cross-modal similarity consistency.
\newblock In {\em Proceedings of the IEEE/CVF Conference on Computer Vision and Pattern Recognition}, pages 19883--19892, 2023.

\bibitem{zhang2025benchmark}
Ruiheng Zhang, Biwen Yang, Lixin Xu, Yan Huang, Xiaofeng Xu, Qi~Zhang, Zhizhuo Jiang, and Yu~Liu.
\newblock A benchmark and frequency compression method for infrared few-shot object detection.
\newblock {\em IEEE Transactions on Geoscience and Remote Sensing}, 2025.

\bibitem{cao2024mentored}
Xiaofeng Cao, Yaming Guo, Heng~Tao Shen, Ivor~W Tsang, and James~T Kwok.
\newblock Mentored learning: Improving generalization and convergence of student learner.
\newblock {\em Journal of Machine Learning Research}, 25(325):1--45, 2024.

\bibitem{huang2025diffusion}
Rui Huang, Shitong Shao, Zikai Zhou, Pukun Zhao, Hangyu Guo, Tian Ye, Lichen Bai, Shuo Yang, and Zeke Xie.
\newblock Diffusion dataset condensation: Training your diffusion model faster with less data.
\newblock {\em arXiv preprint arXiv:2507.05914}, 2025.

\bibitem{zhang2021deep}
Ruiheng Zhang, Lixin Xu, Zhengyu Yu, Ye~Shi, Chengpo Mu, and Min Xu.
\newblock Deep-irtarget: An automatic target detector in infrared imagery using dual-domain feature extraction and allocation.
\newblock {\em IEEE Transactions on Multimedia}, 24:1735--1749, 2021.

\bibitem{yang2022estimating}
Shuo Yang, Erkun Yang, Bo~Han, Yang Liu, Min Xu, Gang Niu, and Tongliang Liu.
\newblock Estimating instance-dependent bayes-label transition matrix using a deep neural network.
\newblock In {\em International Conference on Machine Learning}, pages 25302--25312. PMLR, 2022.

\bibitem{koh2017understanding}
Pang~Wei Koh and Percy Liang.
\newblock Understanding black-box predictions via influence functions.
\newblock In {\em International conference on machine learning}, pages 1885--1894. PMLR, 2017.

\bibitem{ghorbani2019data}
Amirata Ghorbani and James Zou.
\newblock Data shapley: Equitable valuation of data for machine learning.
\newblock In {\em International conference on machine learning}, pages 2242--2251. PMLR, 2019.

\bibitem{kwon2021efficient}
Yongchan Kwon, Manuel~A Rivas, and James Zou.
\newblock Efficient computation and analysis of distributional shapley values.
\newblock In {\em International Conference on Artificial Intelligence and Statistics}, pages 793--801. PMLR, 2021.

\bibitem{jiang2020characterizing}
Ziheng Jiang, Chiyuan Zhang, Kunal Talwar, and Michael~C Mozer.
\newblock Characterizing structural regularities of labeled data in overparameterized models.
\newblock {\em arXiv preprint arXiv:2002.03206}, 2020.

\bibitem{nohyun2022data}
Ki~Nohyun, Hoyong Choi, and Hye~Won Chung.
\newblock Data valuation without training of a model.
\newblock In {\em The Eleventh International Conference on Learning Representations}, 2022.

\bibitem{wu2022davinz}
Zhaoxuan Wu, Yao Shu, and Bryan Kian~Hsiang Low.
\newblock Davinz: Data valuation using deep neural networks at initialization.
\newblock In {\em International Conference on Machine Learning}, pages 24150--24176. PMLR, 2022.

\bibitem{zhang2022rethinking}
Rui Zhang and Shihua Zhang.
\newblock Rethinking influence functions of neural networks in the over-parameterized regime.
\newblock In {\em Proceedings of the AAAI Conference on Artificial Intelligence}, volume~36, pages 9082--9090, 2022.

\bibitem{yang2022dataset}
Shuo Yang, Zeke Xie, Hanyu Peng, Min Xu, Mingming Sun, and Ping Li.
\newblock Dataset pruning: Reducing training data by examining generalization influence.
\newblock {\em arXiv preprint arXiv:2205.09329}, 2022.

\bibitem{grabisch1999axiomatic}
Michel Grabisch and Marc Roubens.
\newblock An axiomatic approach to the concept of interaction among players in cooperative games.
\newblock {\em International Journal of game theory}, 28:547--565, 1999.

\bibitem{bordt2023shapley}
Sebastian Bordt and Ulrike von Luxburg.
\newblock From shapley values to generalized additive models and back.
\newblock In {\em International Conference on Artificial Intelligence and Statistics}, pages 709--745. PMLR, 2023.

\bibitem{yeh2018representer}
Chih-Kuan Yeh, Joon Kim, Ian En-Hsu Yen, and Pradeep~K Ravikumar.
\newblock Representer point selection for explaining deep neural networks.
\newblock {\em Advances in neural information processing systems}, 31, 2018.

\bibitem{pruthi2020estimating}
Garima Pruthi, Frederick Liu, Satyen Kale, and Mukund Sundararajan.
\newblock Estimating training data influence by tracing gradient descent.
\newblock {\em Advances in Neural Information Processing Systems}, 33:19920--19930, 2020.

\bibitem{chen2021hydra}
Yuanyuan Chen, Boyang Li, Han Yu, Pengcheng Wu, and Chunyan Miao.
\newblock Hydra: Hypergradient data relevance analysis for interpreting deep neural networks.
\newblock In {\em Proceedings of the AAAI Conference on Artificial Intelligence}, volume~35, pages 7081--7089, 2021.

\bibitem{hammoudeh2024training}
Zayd Hammoudeh and Daniel Lowd.
\newblock Training data influence analysis and estimation: A survey.
\newblock {\em Machine Learning}, 113(5):2351--2403, 2024.

\bibitem{zhu2016we}
Xiangxin Zhu, Carl Vondrick, Charless~C Fowlkes, and Deva Ramanan.
\newblock Do we need more training data?
\newblock {\em International Journal of Computer Vision}, 119(1):76--92, 2016.

\bibitem{zhang2025logosp}
Zihui Zhang, Weisheng Dai, Hongtao Wen, and Bo~Yang.
\newblock Logosp: Local-global grouping of superpoints for unsupervised semantic segmentation of 3d point clouds.
\newblock In {\em Proceedings of the Computer Vision and Pattern Recognition Conference}, pages 1374--1384, 2025.

\bibitem{liu2025trackvla++}
Jiahang Liu, Yunpeng Qi, Jiazhao Zhang, Minghan Li, Shaoan Wang, Kui Wu, Hanjing Ye, Hong Zhang, Zhibo Chen, Fangwei Zhong, et~al.
\newblock Trackvla++: Unleashing reasoning and memory capabilities in vla models for embodied visual tracking.
\newblock {\em arXiv preprint arXiv:2510.07134}, 2025.

\bibitem{zhou2024few}
Yiwei Zhou, Xiaobo Xia, Zhiwei Lin, Bo~Han, and Tongliang Liu.
\newblock Few-shot adversarial prompt learning on vision-language models.
\newblock {\em Advances in Neural Information Processing Systems}, 37:3122--3156, 2024.

\bibitem{zhang2025visible}
Ruiheng Zhang, Zhe Cao, Yan Huang, Shuo Yang, Lixin Xu, and Min Xu.
\newblock Visible-infrared person re-identification with real-world label noise.
\newblock {\em IEEE Transactions on Circuits and Systems for Video Technology}, 2025.

\bibitem{huang2022harnessing}
Zhuo Huang, Xiaobo Xia, Li~Shen, Bo~Han, Mingming Gong, Chen Gong, and Tongliang Liu.
\newblock Harnessing out-of-distribution examples via augmenting content and style.
\newblock {\em arXiv preprint arXiv:2207.03162}, 2022.

\bibitem{dunlap2023diversify}
Lisa Dunlap, Alyssa Umino, Han Zhang, Jiezhi Yang, Joseph~E Gonzalez, and Trevor Darrell.
\newblock Diversify your vision datasets with automatic diffusion-based augmentation.
\newblock {\em Advances in neural information processing systems}, 36:79024--79034, 2023.

\bibitem{liao2025bood}
Qilin Liao, Shuo Yang, Bo~Zhao, Ping Luo, and Hengshuang Zhao.
\newblock Bood: Boundary-based out-of-distribution data generation.
\newblock {\em arXiv preprint arXiv:2508.00350}, 2025.

\bibitem{yang2021free}
Shuo Yang, Lu~Liu, and Min Xu.
\newblock Free lunch for few-shot learning: Distribution calibration.
\newblock {\em arXiv preprint arXiv:2101.06395}, 2021.

\bibitem{hendrycks2019augmix}
Dan Hendrycks, Norman Mu, Ekin~D Cubuk, Barret Zoph, Justin Gilmer, and Balaji Lakshminarayanan.
\newblock Augmix: A simple data processing method to improve robustness and uncertainty.
\newblock {\em arXiv preprint arXiv:1912.02781}, 2019.

\bibitem{zhang2020does}
Linjun Zhang, Zhun Deng, Kenji Kawaguchi, Amirata Ghorbani, and James Zou.
\newblock How does mixup help with robustness and generalization?
\newblock {\em arXiv preprint arXiv:2010.04819}, 2020.

\bibitem{zhong2020random}
Zhun Zhong, Liang Zheng, Guoliang Kang, Shaozi Li, and Yi~Yang.
\newblock Random erasing data augmentation.
\newblock In {\em Proceedings of the AAAI conference on artificial intelligence}, volume~34, pages 13001--13008, 2020.

\bibitem{devries2017improved}
Terrance DeVries and Graham~W Taylor.
\newblock Improved regularization of convolutional neural networks with cutout.
\newblock {\em arXiv preprint arXiv:1708.04552}, 2017.

\bibitem{ciocca2007self}
Gianluigi Ciocca, Claudio Cusano, Francesca Gasparini, and Raimondo Schettini.
\newblock Self-adaptive image cropping for small displays.
\newblock {\em IEEE Transactions on Consumer Electronics}, 53(4):1622--1627, 2007.

\bibitem{goodfellow2014generative}
Ian~J Goodfellow, Jean Pouget-Abadie, Mehdi Mirza, Bing Xu, David Warde-Farley, Sherjil Ozair, Aaron Courville, and Yoshua Bengio.
\newblock Generative adversarial nets.
\newblock {\em Advances in neural information processing systems}, 27, 2014.

\bibitem{frid2018synthetic}
Maayan Frid-Adar, Eyal Klang, Michal Amitai, Jacob Goldberger, and Hayit Greenspan.
\newblock Synthetic data augmentation using gan for improved liver lesion classification.
\newblock In {\em 2018 IEEE 15th international symposium on biomedical imaging (ISBI 2018)}, pages 289--293. IEEE, 2018.

\bibitem{zhu2017unpaired}
Jun-Yan Zhu, Taesung Park, Phillip Isola, and Alexei~A Efros.
\newblock Unpaired image-to-image translation using cycle-consistent adversarial networks.
\newblock In {\em Proceedings of the IEEE international conference on computer vision}, pages 2223--2232, 2017.

\bibitem{nichol2021glide}
Alex Nichol, Prafulla Dhariwal, Aditya Ramesh, Pranav Shyam, Pamela Mishkin, Bob McGrew, Ilya Sutskever, and Mark Chen.
\newblock Glide: Towards photorealistic image generation and editing with text-guided diffusion models.
\newblock {\em arXiv preprint arXiv:2112.10741}, 2021.

\bibitem{zhu2024distribution}
Haowei Zhu, Ling Yang, Jun-Hai Yong, Hongzhi Yin, Jiawei Jiang, Meng Xiao, Wentao Zhang, and Bin Wang.
\newblock Distribution-aware data expansion with diffusion models.
\newblock {\em arXiv preprint arXiv:2403.06741}, 2024.

\bibitem{gal2022image}
Rinon Gal, Yuval Alaluf, Yuval Atzmon, Or~Patashnik, Amit~H Bermano, Gal Chechik, and Daniel Cohen-Or.
\newblock An image is worth one word: Personalizing text-to-image generation using textual inversion.
\newblock {\em arXiv preprint arXiv:2208.01618}, 2022.

\bibitem{wallace2024diffusion}
Bram Wallace, Meihua Dang, Rafael Rafailov, Linqi Zhou, Aaron Lou, Senthil Purushwalkam, Stefano Ermon, Caiming Xiong, Shafiq Joty, and Nikhil Naik.
\newblock Diffusion model alignment using direct preference optimization.
\newblock In {\em Proceedings of the IEEE/CVF Conference on Computer Vision and Pattern Recognition}, pages 8228--8238, 2024.

\bibitem{kingma2021variational}
Diederik Kingma, Tim Salimans, Ben Poole, and Jonathan Ho.
\newblock Variational diffusion models.
\newblock {\em Advances in neural information processing systems}, 34:21696--21707, 2021.

\bibitem{nilsback2008automated}
Maria-Elena Nilsback and Andrew Zisserman.
\newblock Automated flower classification over a large number of classes.
\newblock In {\em 2008 Sixth Indian conference on computer vision, graphics \& image processing}, pages 722--729. IEEE, 2008.

\bibitem{zhou2024golden}
Zikai Zhou, Shitong Shao, Lichen Bai, Zhiqiang Xu, Bo~Han, and Zeke Xie.
\newblock Golden noise for diffusion models: A learning framework.
\newblock {\em arXiv preprint arXiv:2411.09502}, 2024.

\bibitem{bai2024zigzag}
Lichen Bai, Shitong Shao, Zikai Zhou, Zipeng Qi, Zhiqiang Xu, Haoyi Xiong, and Zeke Xie.
\newblock Zigzag diffusion sampling: Diffusion models can self-improve via self-reflection.
\newblock In {\em The Thirteenth International Conference on Learning Representations}, volume~2, 2024.

\bibitem{ahn2024noise}
Donghoon Ahn, Jiwon Kang, Sanghyun Lee, Jaewon Min, Minjae Kim, Wooseok Jang, Hyoungwon Cho, Sayak Paul, SeonHwa Kim, Eunju Cha, et~al.
\newblock A noise is worth diffusion guidance.
\newblock {\em arXiv preprint arXiv:2412.03895}, 2024.

\bibitem{yeo2024controlled}
Teresa Yeo, Andrei Atanov, Harold Benoit, Aleksandr Alekseev, Ruchira Ray, Pooya~Esmaeil Akhoondi, and Amir Zamir.
\newblock Controlled training data generation with diffusion models.
\newblock {\em arXiv preprint arXiv:2403.15309}, 2024.

\bibitem{deng2009imagenet}
Jia Deng, Wei Dong, Richard Socher, Li-Jia Li, Kai Li, and Li~Fei-Fei.
\newblock Imagenet: A large-scale hierarchical image database.
\newblock In {\em 2009 IEEE conference on computer vision and pattern recognition}, pages 248--255. Ieee, 2009.

\bibitem{fei2004learning}
Li~Fei-Fei, Rob Fergus, and Pietro Perona.
\newblock Learning generative visual models from few training examples: An incremental bayesian approach tested on 101 object categories.
\newblock In {\em 2004 conference on computer vision and pattern recognition workshop}, pages 178--178. IEEE, 2004.

\bibitem{parkhi2012cats}
Omkar~M Parkhi, Andrea Vedaldi, Andrew Zisserman, and CV~Jawahar.
\newblock Cats and dogs.
\newblock In {\em 2012 IEEE conference on computer vision and pattern recognition}, pages 3498--3505. IEEE, 2012.

\bibitem{bossard2014food}
Lukas Bossard, Matthieu Guillaumin, and Luc Van~Gool.
\newblock Food-101--mining discriminative components with random forests.
\newblock In {\em Computer vision--ECCV 2014: 13th European conference, zurich, Switzerland, September 6-12, 2014, proceedings, part VI 13}, pages 446--461. Springer, 2014.

\bibitem{wah2011caltech}
Catherine Wah, Steve Branson, Peter Welinder, Pietro Perona, and Serge Belongie.
\newblock The caltech-ucsd birds-200-2011 dataset.
\newblock 2011.

\bibitem{cimpoi2014describing}
Mircea Cimpoi, Subhransu Maji, Iasonas Kokkinos, Sammy Mohamed, and Andrea Vedaldi.
\newblock Describing textures in the wild.
\newblock In {\em Proceedings of the IEEE conference on computer vision and pattern recognition}, pages 3606--3613, 2014.

\bibitem{guo2025deepseek}
Daya Guo, Dejian Yang, Haowei Zhang, Junxiao Song, Ruoyu Zhang, Runxin Xu, Qihao Zhu, Shirong Ma, Peiyi Wang, Xiao Bi, et~al.
\newblock Deepseek-r1: Incentivizing reasoning capability in llms via reinforcement learning.
\newblock {\em arXiv preprint arXiv:2501.12948}, 2025.

\end{thebibliography}

\clearpage
\newpage
\section*{NeurIPS Paper Checklist}

\begin{enumerate}

\item {\bf Claims}
    \item[] Question: Do the main claims made in the abstract and introduction accurately reflect the paper's contributions and scope?
    \item[] Answer: \answerYes{} 
    \item[] Justification: We confirm that the abstract and introduction accurately reflects the paper's contributions, as it clearly outlines the proposed \textsc{UtilGen} framework, its adaptive optimization approach, and experimental validation across diverse datasets.

\item {\bf Limitations}
    \item[] Question: Does the paper discuss the limitations of the work performed by the authors?
    \item[] Answer: \answerYes{}
    \item[] Justification:  Discussed in  Appendix \ref{appendix:limitations}

\item {\bf Theory assumptions and proofs}
    \item[] Question: For each theoretical result, does the paper provide the full set of assumptions and a complete (and correct) proof?
    \item[] Answer: \answerNA{}
    \item[] Justification: This paper does not include theoretical results. 

    \item {\bf Experimental result reproducibility}
    \item[] Question: Does the paper fully disclose all the information needed to reproduce the main experimental results of the paper to the extent that it affects the main claims and/or conclusions of the paper (regardless of whether the code and data are provided or not)?
    \item[] Answer: \answerYes{}
    \item[] Justification: We provide comprehensive implementation details in Appendix~\ref{imp_detail} and dataset specifications in Appendix~\ref{tab:dataset_detail}, including methodological implementations and dataset characteristics, to ensure full reproducibility of our work.

\item {\bf Open access to data and code}
    \item[] Question: Does the paper provide open access to the data and code, with sufficient instructions to faithfully reproduce the main experimental results, as described in supplemental material?
    \item[] Answer: \answerYes{}
    \item[] Justification: Our study uses exclusively publicly accessible datasets and includes the complete implementation source code in the supplementary materials (provided as a ZIP archive).

\item {\bf Experimental setting/details}
    \item[] Question: Does the paper specify all the training and test details (e.g., data splits, hyperparameters, how they were chosen, type of optimizer, etc.) necessary to understand the results?
    \item[] Answer: \answerYes{}
    \item[] Justification: We provide comprehensive implementation details, including data partitioning, hyperparameter settings, and additional relevant information in Appendix~\ref{imp_detail} and Appendix~\ref{tab:dataset_detail}.

\item {\bf Experiment statistical significance}
    \item[] Question: Does the paper report error bars suitably and correctly defined or other appropriate information about the statistical significance of the experiments?
    \item[] Answer: \answerNo{}
    \item[] Justification: We opted not to report error bars because the results demonstrated consistent stability across multiple runs, reducing the necessity for such reporting. To ensure robustness, each experiment was conducted using three different random seeds with results averaged accordingly. Additionally, extensive evaluations across diverse datasets further strengthen the reliability of our findings.

\item {\bf Experiments compute resources}
    \item[] Question: For each experiment, does the paper provide sufficient information on the computer resources (type of compute workers, memory, time of execution) needed to reproduce the experiments?
    \item[] Answer: \answerYes{}
    \item[] Justification: We provide detailed information on the computational resources used in this work, including the types and numbers of GPUs, GPU memory usage during execution, and overall running time. For more details, please refer to Appendix~\ref{appendix:scaling}.
    
\item {\bf Code of ethics}
    \item[] Question: Does the research conducted in the paper conform, in every respect, with the NeurIPS Code of Ethics \url{https://neurips.cc/public/EthicsGuidelines}?
    \item[] Answer: \answerYes{}
    \item[] Justification: We have carefully examined the NeurIPS Code of Ethics and affirm that all aspects of this research comply with its guidelines.

\item {\bf Broader impacts}
    \item[] Question: Does the paper discuss both potential positive societal impacts and negative societal impacts of the work performed?
    \item[] Answer: \answerYes{}
    \item[] Justification: We provide a discussion of both positive and negative societal implications in Appendix~\ref{Broader Impact}.
    
\item {\bf Safeguards}
    \item[] Question: Does the paper describe safeguards that have been put in place for responsible release of data or models that have a high risk for misuse (e.g., pretrained language models, image generators, or scraped datasets)?
    \item[] Answer: \answerNA{}
    \item[] Justification: This research does not involve releasing new models or datasets, thus the discussion of release safeguards is not applicable to our work.

\item {\bf Licenses for existing assets}
    \item[] Question: Are the creators or original owners of assets (e.g., code, data, models), used in the paper, properly credited and are the license and terms of use explicitly mentioned and properly respected?
    \item[] Answer: \answerYes{} 
    \item[] Justification: All datasets (e.g., ImageNet \cite{deng2009imagenet}) and foundation models (e.g., Stable Diffusion \cite{rombach2022high}) used in this work are appropriately cited the paper.

\item {\bf New assets}
    \item[] Question: Are new assets introduced in the paper well documented and is the documentation provided alongside the assets?
    \item[] Answer: \answerYes{}
    \item[] Justification: The code documentation is provided in the supplementary materials.

\item {\bf Crowdsourcing and research with human subjects}
    \item[] Question: For crowdsourcing experiments and research with human subjects, does the paper include the full text of instructions given to participants and screenshots, if applicable, as well as details about compensation (if any)? 
    \item[] Answer: \answerNA{}
    \item[] Justification: The paper does not involve crowdsourcing nor research with human subjects.

\item {\bf Institutional review board (IRB) approvals or equivalent for research with human subjects}
    \item[] Question: Does the paper describe potential risks incurred by study participants, whether such risks were disclosed to the subjects, and whether Institutional Review Board (IRB) approvals (or an equivalent approval/review based on the requirements of your country or institution) were obtained?
    \item[] Answer: \answerNA{}
    \item[] Justification: The paper does not involve crowdsourcing nor research with human subjects.

\item {\bf Declaration of LLM usage}
    \item[] Question: Does the paper describe the usage of LLMs if it is an important, original, or non-standard component of the core methods in this research? Note that if the LLM is used only for writing, editing, or formatting purposes and does not impact the core methodology, scientific rigorousness, or originality of the research, declaration is not required.
    \item[] Answer: \answerYes{} 
    \item[] Justification: LLMs were used  for language polishing (grammar and spelling corrections).

\end{enumerate}


\appendix
\newpage

\section{Additional Visualizations of Synthetic Data}
\label{appendix:visualization}

\begin{figure}[h] 
    \centering
    \includegraphics[width=\textwidth]{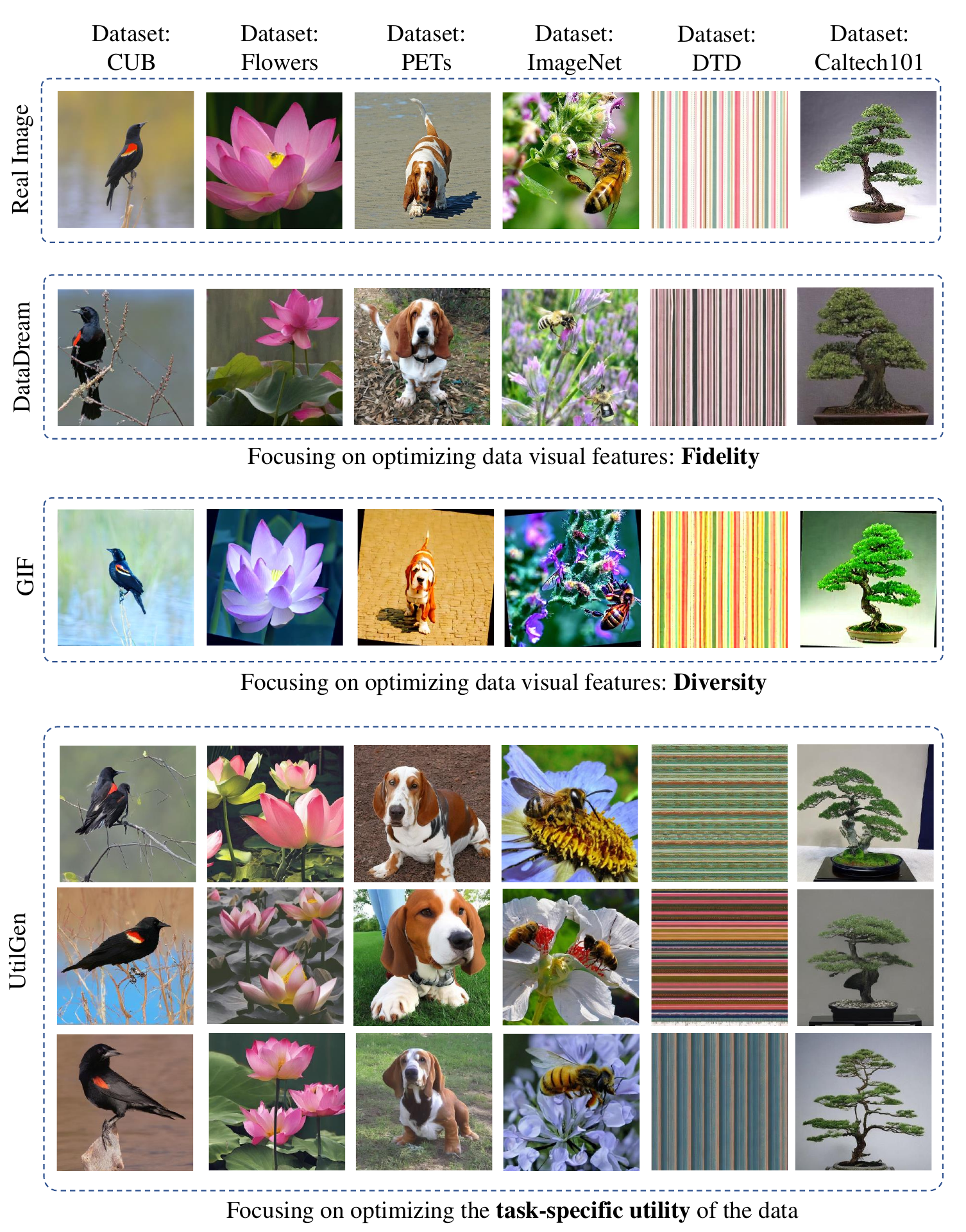}
    \caption{Comparison of synthetic samples generated by GIF, DataDream, and \textsc{UtilGen}.}
    \label{fig:visual2} 
\end{figure}

In this section, we present further visualizations of synthetic data generated by the \textsc{UtilGen}. As illustrated in Figure \ref{fig:visual2}, we compare synthetic samples produced by different augmentation strategies, including GIF (prioritizing diversity in visual features) and DataDream (emphasizing fidelity in visual features). Figure \ref{fig:visual2} presents representative samples from \textsc{UtilGen}, demonstrating semantically consistent, visually realistic generations that preserve class-discriminative features while maintaining diversity.

\section{More Implementation Details}

\label{imp_detail}

\subsection{Implementation Details of Weight Network}
\label{sec:weight_net_impl}

The weight network $\mathcal{W}_\phi$ is designed as a lightweight MLP with a single hidden layer. It takes the per-sample classification loss $\ell_i = \mathcal{L}(f(x_i;\theta), y_i)$ as input, where $f(\cdot;\theta)$ is the downstream classifier, and predicts normalized sample weights $\omega_i \in [0,1]$ via a sigmoid activation. These weights are used to re-weight the corresponding per-sample losses during classifier training, thereby prioritizing examples deemed to have higher utility by the meta-learned network. The network architecture is defined as:
\begin{equation}
\mathcal{W}_\phi(\ell_i) = \sigma(W_2 \, \text{ReLU}(W_1 \ell_i + b_1) + b_2)
\end{equation}
where $\ell_i \in \mathbb{R}$ is the scalar loss value for the $i$-th sample, $W_1 \in \mathbb{R}^{1 \times 100}$ and $W_2 \in \mathbb{R}^{100 \times 1}$ are weight matrices, and $b_1$, $b_2$ are bias terms. The ReLU activation introduces non-linearity, and the final sigmoid ensures the output lies in $[0,1]$. Albeit simple, this network is a universal approximator for continuous functions and can fit a wide range of weighting functions. The training settings for both the classifier and the weight network, including optimizers, learning rates, and batch sizes, are summarized in Table~\ref{tab:params}.

\begin{table}[h]
\centering
\caption{Training settings for the classifier and weight network.}
\label{tab:params}
\begin{tabular}{lccc}
\toprule
Component & Optimizer & Learning Rate & Batch Size \\
\midrule
Classifier & SGD (momentum=0.9) & 0.01 & 128 \\
Weight Network & Adam & $10^{-3}$ & 128 \\
\bottomrule
\end{tabular}
\end{table}

\subsection{Implementation Details of MLCO}

\begin{algorithm}
\caption{Model-Level Generation Capability Optimization (MLCO)}\label{algo_mlco}
\footnotesize
\begin{flushleft}
\textbf{Input}: Initial diffusion model $g_\psi$; class prompts $\{c_1,...,c_K\}$.

\textbf{Required}: Batch size $B$; DPO learning rate $\eta$; reference model $g_{\text{ref}}$; weight network $\mathcal{W}_\phi$; downstream classifier $f_\theta$; loss function $\mathcal{L}$; max iteration $I$; selection ratio $\rho$..
\end{flushleft}

\begin{algorithmic}[1]
\For{iteration = 1 to $I$}
    \State \textbf{// Generation \& Evaluation}
    \For{each class $k \in \{1,...,K\}$}
        \For{$i = 1$ to $B$}
            \State Sample $\epsilon_T \sim \mathcal{N}(0,I)$
            \State $x_i^k \gets g_\psi(c_k, \epsilon_T)$ \Comment{Generate with current model}
            \State $\omega_i^k \gets \mathcal{W}_\phi(\mathcal{L}(f(x_i^k;\theta), y_k))$ \Comment{Evaluate utility using classification loss as input}
        \EndFor
    \EndFor
    
    \State \textbf{// Preference Construction}
    \For{each class $k$}
        \State Sort $\{x_i^k\}$ by $\omega_i^k$ descending
        \State Select top $\rho$ proportion of samples as $\mathcal{D}_k^w$, bottom $\rho$ proportion of samples as $\mathcal{D}_k^l$
        \State $\mathcal{D}_{\text{pref}} \gets \mathcal{D}_{\text{pref}} \cup \{(c_k, x^w, x^l) | x^w \in \mathcal{D}_k^w, x^l \in \mathcal{D}_k^l\}$
    \EndFor
    
    \State \textbf{// Model Optimization}
    \For{each $(c_k, x^w, x^l)$ in $\mathcal{D}_{\text{pref}}$}
        \State $\psi \gets \psi - \eta \nabla_{\psi}\mathcal{L}_{\text{DPO}}(g_\psi, g_{\text{ref}}, c_k, x^w, x^l)$ \Comment{Update via Eq.~\ref{eq:dpo_loss}}
    \EndFor
\EndFor
\end{algorithmic}
\begin{flushleft}
\textbf{Output}: Optimized diffusion model $g_\psi^*$
\end{flushleft}
\end{algorithm}

The proposed Model-Level Generation Capability Optimization (MLCO) framework iteratively improves the diffusion model’s generative ability based on utility-guided preferences. In each iteration, the process proceeds in three stages, and he full process is illustrated in Algorithm~\ref{algo_mlco}.:

\begin{itemize}[leftmargin=*]
    \item \textbf{Generation \& Evaluation:} The current diffusion model $g_\psi$ generates synthetic images for each class prompt. Each generated sample is then evaluated by the trained weight network $\mathcal{W}_\phi$ to obtain utility scores $\omega$.
    \item \textbf{Preference Construction:} Samples within each class are ranked by their utility scores. The top $\rho$ and bottom $\rho$ proportions are selected to form preference pairs $\mathcal{D}_{\text{pref}}$.
    \item \textbf{Model Optimization:} Using Direct Preference Optimization (DPO), the model is updated to customize its generation capability based on utility-guided preferences over training data.
\end{itemize}

Our implementation of DPO is adapted from Diffusion-DPO~\cite{wallace2024diffusion}, which enables preference-based training of diffusion models for guiding generative outputs with utility-informed preferences. The loss function Eq.~\ref{eq:dpo_loss} used in Line~19 of Algorithm~\ref{algo_mlco} follows this framework. For the detailed mathematical derivation of Eq.~\ref{eq:dpo_loss}, please refer to the original Diffusion-DPO paper~\cite{wallace2024diffusion}. The key hyperparameters used in our DPO training process are summarized in Table~\ref{tab:dpo_hyperparams}.

\begin{table}[H]
\renewcommand{\arraystretch}{1.5}
\centering
\caption{Hyperparameters used in DPO training}
\label{tab:dpo_hyperparams}
\begin{tabular}{ccccc}
\hline
Batch Size & Max Steps Per Class& Learning Rate & Gradient Accumulation Steps & Beta DPO \\
\hline
1 & 400 & $1 \times 10^{-8}$ & 1 & 5000 \\
\hline
\end{tabular}
\end{table}

\subsection{Implementation Details of ILPO}

The prompt-noise optimization process consists of two primary components: the optimization of prompt embeddings and the initial noise used during generation. The goal is to maximize the utility of each synthetic sample while ensuring semantic alignment with the target domain.

\textbf{Prompt embedding optimization:} The optimization process begins with textual inversion \cite{gal2022image} to establish class-specific prompt embeddings that align with target concepts. Specifically, we use DeepSeek-R1-Distill-Qwen-1.5B \cite{guo2025deepseek} to select an initializer token for textual inversion \cite{gal2022image} for each class. After obtaining the class-specific prompt embeddings aligned with the target labels, these embeddings are refined through gradient-based optimization to maximize the utility score of the generated samples. The learning rate for prompt optimization is set to 0.001, and the optimization process runs for 400 epochs to ensure convergence and meaningful results.

\begin{table}[H]
\centering
\caption{Hyperparameters used in textual inversion \cite{gal2022image}}
\renewcommand{\arraystretch}{1.5}
\label{tab:textual_inversion_params}
\begin{tabular}{cccc}
\hline
Batch Size & Learning Rate & Training Steps & Instance Images per Class  \\
\hline
1 & $1\times10^{-4}$ & 400 & 16 \\
\hline
\end{tabular}
\end{table}

\textbf{Noise optimization:} For noise optimization, we leverage the discrepancy between denoising and inversion Classifier-Free Guidance (CFG) scales, which allows us to implicitly inject prompt semantic information into the noise vector. The denoising guidance scale is set to 5.5, while the inversion guidance scale is set to 0. 

The specific hyperparameters used for this optimization process are summarized in Table~\ref{tab:implementation_details}.

\begin{table}[H]
\renewcommand{\arraystretch}{1.3}
\centering
\caption{Hyperparameters used in the prompt-noise optimization process.}
\label{tab:implementation_details}
\begin{tabular}{cccc}
\hline
\multicolumn{1}{c}{\multirow{2}{*}{Prompt Learning Rate}} & 
\multicolumn{1}{c}{\multirow{2}{*}{Prompt Learning Epochs}} & 
\multicolumn{1}{c}{\multirow{2}{*}{\makecell{Guidance Strength \\ (Denoise)}}} & 
\multicolumn{1}{c}{\multirow{2}{*}{\makecell{Guidance Strength \\ (Inversion)}}} \\ 
& & & \\ \hline
0.001 & 400 & 5.5 & 0 \\ \hline
\end{tabular}
\end{table}

\subsection{Implementation Details of Image Generation}

The image generation process is guided by the hyperparameters listed in Table~\ref{tab:hyperparams}. Instead of using fixed prompts and random noise, both the class-specific text prompts and the initial noise vectors are optimized via our proposed ILPO strategy to better align with high-utility regions in the data distribution. The Stable Diffusion v2.1 model, fine-tuned with MLCO, is employed for the generation process, using a sampling method with 50 steps, the DDIM scheduler, and a guidance scale of 2.0. The images are generated at a resolution of $512 \times 512$ pixels.

\begin{table}[H]
\caption{Hyperparameters for Training Data Synthesis}
\label{tab:hyperparams}
\centering
\begin{tabular}{lcccc}
\toprule
Base Model & Sampling Steps & Scheduler & Guidance Scale & Image Size \\
\midrule
\multicolumn{1}{c}{\multirow{2}{*}{\makecell{Stable Diffusion v2.1\\(MLCO-fine-tuned)}}} & \multicolumn{1}{c}{\multirow{2}{*}{50}} & \multicolumn{1}{c}{\multirow{2}{*}{DDIM}} & \multicolumn{1}{c}{\multirow{2}{*}{2.0}} & \multicolumn{1}{c}{\multirow{2}{*}{512×512}} \\
&&&&\\
\bottomrule
\end{tabular}
\end{table}

\subsection{Implementation details of model training}

The downstream classifiers are trained on four standard architectures: ResNet-50\cite{he2016deep}, ResNeXt-50 \cite{xie2017aggregated}, WideResNet-50 \cite{zagoruyko2016wide}, and MobileNetV2 \cite{sandler2018mobilenetv2}. All models are trained with identical hyperparameters. The training configuration uses SGD optimizer with momentum 0.9 and weight decay 5e-4. The learning rate starts at 0.01 with cosine decay schedule over 100 epochs. A fixed batch size of 256 is used for all experiments, with standard data augmentation including random horizontal flips and crops. Each experiment is repeated three times with different random seeds to ensure reliability.

\begin{table}[H]
\centering
\caption{Hyperparameters for Downstream Classifier Training}
\renewcommand{\arraystretch}{1.2}
\label{tab:training_params}
\begin{tabular}{lccccc}
\toprule
Optimizer & Weight decay & Initial LR  & Epochs & Batch size  \\
\midrule
SGD (momentum=0.9) & 5e-4 & 0.01  & 100 & 256  \\
\bottomrule
\end{tabular}
\end{table}

\section{Limitations}
\label{appendix:limitations}
While demonstrating strong performance in enhancing synthetic data utility for downstream tasks, \textsc{UtilGen} presents two noteworthy considerations: (1) The dual-level optimization framework incurs modest computational overhead compared to conventional augmentation methods; (2) Although effectively improving the utility of synthetic data for downstream tasks, the approach remains contingent upon the base generative model's capability to produce viable initial samples. These considerations do not substantially compromise overall performance but indicate potential avenues for future enhancement.

\section{Efficiency and Cost of Data Augmentation}
\label{appendix:scaling}
To evaluate the computational efficiency of \textsc{UtilGen}, we compare it against three representative generative augmentation methods: GIF \cite{zhang2023expanding}, GAP \cite{yeo2024controlled}, and DataDream \cite{kim2024datadream}. The comparison is conducted under a unified setup using the ImageNet-1K dataset, where each method generates 1000 synthetic images per class. We decompose the computational pipeline into three stages: (1) \textit{Model Optimization}, (2) \textit{Policy Optimization}, and (3) \textit{Image Generation}. \textsc{UtilGen} incorporates feedback-driven optimization components at both the model and instance levels, introducing moderate overhead that remains manageable.

Table~\ref{tab:comp_cost} presents the runtime taken for each stage, while Table~\ref{tab:memory_req} shows the peak GPU memory usage during the execution of each stage. All methods are evaluated on a multi-GPU server equipped with 8×A100 GPUs. Despite the feedback-based optimization, our framework remains computationally efficient. Overall, \textsc{UtilGen} strikes a favorable balance between computational cost and data utility, demonstrating its scalability and practicality for real-world deployments.

\begin{table}[h]
\centering
\caption{Computational cost (in hours) comparison on ImageNet-1K for generating 1000 images per class. Values are estimated averages.}
\label{tab:comp_cost}
\renewcommand{\arraystretch}{1.2}
\setlength{\tabcolsep}{6pt}
\begin{tabular}{lccc}
\toprule
Method & Model Optimization & Policy Optimization & Image Generation \\
\midrule
GIF \cite{zhang2023expanding} & -- & -- & $\sim$229.1h \\
GAP \cite{yeo2024controlled} & -- & $\sim$45.1h & $\sim$93.7h \\
DataDream \cite{kim2024datadream} & $\sim$10.8h & -- & $\sim$40.2h \\
\textsc{UtilGen} (Ours) & $\sim$7.7h & $\sim$25.1h & $\sim$41.6h \\
\bottomrule
\end{tabular}
\end{table}

\begin{table}[h]
\centering
\caption{GPU memory usage (peak usage in GB) per stage. All values are measured during maximum workload per module on each GPU.}
\label{tab:memory_req}
\renewcommand{\arraystretch}{1.2}
\setlength{\tabcolsep}{6pt}
\begin{tabular}{lccc}
\toprule
Method & Model Optimization & Policy Optimization & Image Generation \\
\midrule
GIF \cite{zhang2023expanding} & -- & -- & $\sim$25.9G \\
GAP \cite{yeo2024controlled} & -- & $\sim$5.2G & $\sim$4.5G \\
DataDream \cite{kim2024datadream} & $\sim$19.5G & -- & $\sim$15.8G \\
\textsc{UtilGen} (Ours) & $\sim$20.5G & $\sim$4.4G & $\sim$4.3G \\
\bottomrule
\end{tabular}
\end{table}

\section{Broader Impact}
\label{Broader Impact}

Our utility-driven augmentation approach facilitates more efficient model training while reducing reliance on real data, especially benefiting domains with limited or private datasets. By generating task-specific synthetic training data, it enhances learning efficiency and lowers dependence on large-scale real datasets. Nonetheless, since the synthesis process is guided by a small set of real images, the generated data may inadvertently inherit and amplify biases present in the original samples.

\section{Dataset Details}
\label{tab:dataset_detail}

To evaluate \textsc{UtilGen}'s performance, we utilize eight benchmark datasets spanning a variety of classification tasks: coarse-grained classification, fine-grained classification, and texture classification.

The coarse-grained datasets include ImageNet-1k-Subset~\cite{deng2009imagenet}, ImageNet-100-Subset~\cite{deng2009imagenet} and Caltech 101~\cite{fei2004learning}. ImageNet-1k-Subset~\cite{deng2009imagenet} and ImageNet-100-Subset~\cite{deng2009imagenet}, both randomly sampled with 100 images per class. ImageNet-100-Subset is a subset of 100 animal-related classes from ImageNet-1K. Caltech 101~\cite{fei2004learning} consists of 101 object categories. For fine-grained classification, we use Oxford Pets~\cite{parkhi2012cats}, Food101-Subset~\cite{bossard2014food}, Flowers 102~\cite{nilsback2008automated}, and CUB-200-2011~\cite{wah2011caltech}, with Food101-Subset~\cite{bossard2014food} being a curated subset of Food101~\cite{bossard2014food} containing 101 food categories. Other datasets follow their original training and validation setups.
Texture classification is evaluated using the DTD~\cite{cimpoi2014describing} dataset, which contains 47 texture categories. Detailed dataset statistics are provided in Table~\ref{tab:dataset_stats}, summarizing the number of classes, training samples, and test samples for each dataset. It is important to note that datasets with a higher number of classes or fewer average samples per class present greater challenges in terms of classification and generalization.

\begin{table}[h]
\renewcommand{\arraystretch}{1.3}
\centering
\setlength{\tabcolsep}{0.8mm}
\caption{Statistics of the benchmark datasets}
\label{tab:dataset_stats}
\begin{tabular}{lcccc}
\hline
Dataset & Task Type & Classes & Training Data &Test Data \\
\hline
ImageNet-1k-Subset~\cite{deng2009imagenet} &  Coarse-grained object classification & 1000 & 100,000&50,000 \\
ImageNet-100-Subset~\cite{deng2009imagenet} &  Coarse-grained object classification & 100 & 10,000&5,000 \\
Caltech 101~\cite{fei2004learning} &  Coarse-grained object classification & 101 & 3060&6084 \\
Oxford Pets~\cite{parkhi2012cats}& Fine-grained object classification & 37 & 3680&3669 \\
Food101-Subset~\cite{bossard2014food}& Fine-grained object classification & 101 & 10100 &25,250 \\
Flowers 102~\cite{nilsback2008automated} & Fine-grained object classification & 102 & 6,552 & 818 \\
CUB-200-2011~\cite{wah2011caltech}& Fine-grained object classification & 200 & 5,994 & 5,794 \\
DTD~\cite{cimpoi2014describing}& Texture classification & 47 & 1880&1,880 \\
\hline
\end{tabular}
\end{table}

\end{document}